\documentclass[letterpaper, 10 pt, conference]{ieeeconf}

\IEEEoverridecommandlockouts
\usepackage{amsmath}

\usepackage{amsfonts}
\usepackage{amsthm}
\usepackage{mathtools}
\usepackage{nicefrac}
\usepackage{xcolor}
\usepackage{hyperref}
\hypersetup{hidelinks}
\usepackage{booktabs}
\usepackage{microtype}

\usepackage{tikz}
\usetikzlibrary{arrows,plotmarks,calc,angles,quotes,patterns}
\usetikzlibrary{decorations.pathmorphing,decorations.markings}

\newtheorem{theorem}{Theorem}
\theoremstyle{remark}
\newtheorem{remark}{\normalfont \bfseries Remark}
\newtheorem{lemma}{\normalfont \bfseries Lemma}

\definecolor{mygreen}{rgb}{0.0, 0.5, 0.0}

\DeclarePairedDelimiter\abs{\lvert}{\rvert}

\newcommand\centerofmass{
    \tikz[radius=0.4em] {
        \fill (0,0) -- ++(0.4em,0) arc [start angle=0,end angle=90] -- ++(0,-0.8em) arc [start angle=270, end angle=180];
        \fill [color=white] (0,0) -- ++(0,0.4em) arc [start angle=90,end angle=180] -- ++(0.8em,0) arc [start angle=0, end angle=-90];
        \draw (0,0) circle;
    }
}

\title{\LARGE \bf
Cable Estimation-Based Control for \\ Wire-Borne Underactuated Brachiating Robots: \\ A Combined Direct-Indirect Adaptive Robust Approach
}

\author{Siavash Farzan, Vahid Azimi, Ai-Ping Hu, and Jonathan Rogers%
\thanks{Siavash Farzan is with the Institute for Robotics and Intelligent Machines, Georgia Institute of Technology,
        Atlanta, GA, USA,
        {\tt\footnotesize sfarzan@gatech.edu}}%
\thanks{Vahid Azimi is with the Department of Energy Resources Engineering, Stanford University, Stanford, CA, USA,
        {\tt\footnotesize vazimi@stanford.edu}}%
\thanks{Ai-Ping Hu is with the Georgia Tech Research Institute,
        Atlanta, GA, USA,
        {\tt\footnotesize ai-ping.hu@gtri.gatech.edu}}
\thanks{Jonathan Rogers is with the Guggenheim School of Aerospace
        Engineering, Georgia Institute of Technology,
        Atlanta, GA, USA,
        {\tt\footnotesize jonathan.rogers@ae.gatech.edu}}%
}

\begin{document}

\maketitle
\thispagestyle{empty}
\pagestyle{empty}

\begin{abstract}
In this paper, we present an online adaptive robust control framework for underactuated brachiating robots traversing flexible cables. Since the dynamic model of a flexible body is unknown in practice, we propose an indirect adaptive estimation scheme to approximate the unknown dynamic effects of the flexible cable as an external force with parametric uncertainties. A boundary layer-based sliding mode control is then designed to compensate for the residual unmodeled dynamics and time-varying disturbances, in which the control gain is updated by an auxiliary direct adaptive control mechanism. Stability analysis and derivation of adaptation laws are carried out through a Lyapunov approach, which formally guarantees the stability and tracking performance of the robot-cable system. Simulation experiments and comparison with a baseline controller show that the combined direct-indirect adaptive robust control framework achieves reliable tracking performance and adaptive system identification, enabling the robot to traverse flexible cables in the presence of unmodeled dynamics, parametric uncertainties and unstructured disturbances.\looseness=-1
\end{abstract}

\section{INTRODUCTION AND MOTIVATION}
Despite the fact that brachiating robots have been studied extensively during the past two decades~\cite{SaiAra94,MazAsa09}, they have not yet emerged in real life scenarios. A major challenge in deploying brachiating robots in real life applications comes from the uncertainties and disturbances present in outdoor environments. Moreover, in the current literature, brachiating robots have been researched almost exclusively for rigid bars/supports~\cite{NakKod00,YanCho19},
which are difficult to be constructed in outdoor settings. By contrast, wire traversing robots~\cite{Davies18,NotEge19}
have a better chance of getting deployed in real life applications, as it is relatively easier to install a flexible wire or cable in outdoor environments. The existing infrastructure such as overhead wires for trolley/bus systems or power transmission lines can be also leveraged as a medium for this purpose.\looseness=-1

\begin{figure}[t]
	\centering
	\includegraphics[trim={0bp 180bp 0bp 60bp},clip, width=0.56\columnwidth]{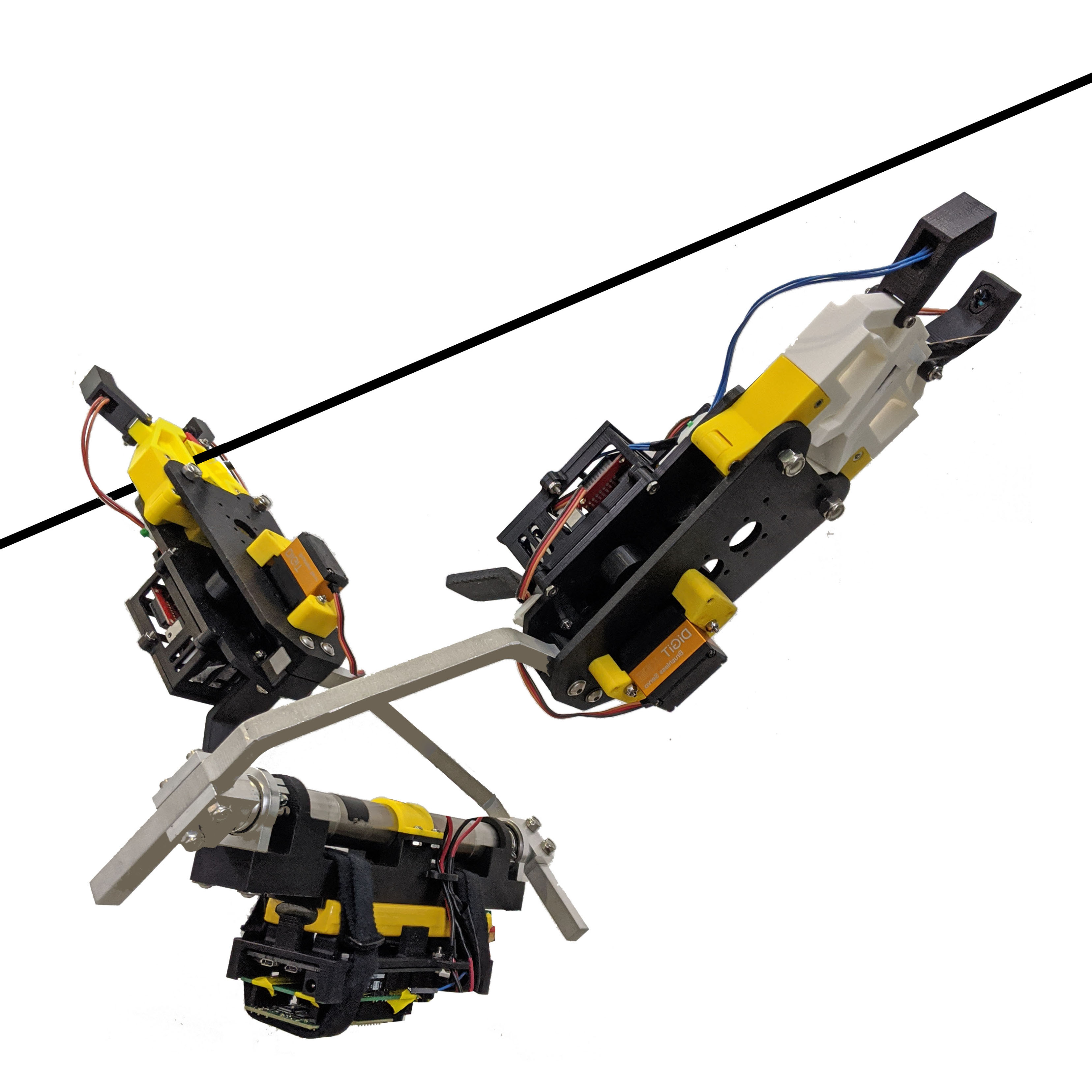}
	\caption{Robot hardware prototype performing a brachiation maneuver.}
	\label{fig:robot-hardware}
\end{figure}

Attaching an underactuated brachiating robot (Fig.~\ref{fig:robot-hardware}) to a flexible cable makes the control task more challenging, as it increases the degrees of underactuation, and introduces unmodeled dynamics and unknown uncertainties/disturbances due to the dynamic effects of an oscillatory cable. Deriving an exact model for a flexible cable is infeasible. Moreover, 
neither the states of a flexible cable nor the force applied by the cable to the robot can be measured in practice using common sensors~\cite{Farzan19}.

To overcome these challenges, the cable dynamics effects and the resulting uncertainties are required to be estimated, and the discrepancies between the approximated and actual model need to be compensated. The former objective motivates the use of an \emph{adaptive} control scheme,
while the latter can be dealt with by using a \emph{robust} control design~\cite{SloCoe86,Farzan20}.

\emph{Indirect} adaptive methods are particularly common in the robotics control literature~\cite{LiSlo89,ShiMos15},
where the adaptive law generates on-line estimates of the unknown parameters of the system dynamics which then are used to calculate the control law. 
However, to guarantee parameter convergence and achieve zero error tracking, adaptive methods rely on the reference trajectory to be persistently exciting~\cite{NarAnn87}, which is not always ensured for dynamical systems. Additionally, the performance of adaptive controllers may be significantly degraded or even lead to instability if disturbances and unmodeled dynamics are too large in the system.

Robust controllers can be employed to mitigate the effects of modeling errors and bounded disturbances on a system's stability and performance. Sliding mode control~\cite{SloWei91}
is an efficient robust control method that has been widely used to control systems with bounded disturbances and uncertainties~\cite{AziSim18},
entailing construction of a surface onto which the error asymptotically converges to zero. However, designing a stable sliding manifold is not straightforward for underactuated systems~\cite{AshErw08}. Moreover, to tune the constant gains of robust control terms, the bounds of modelling errors and disturbances need to be known in advance, which is not the case for many applications. \emph{Direct} adaptive methods~\cite{UlrBar12,AziVel20}
can be applied to form a time-varying control gain and automatically compensate for bounded disturbances without the need to know the bounds a priori. Using a direct adaptive design, instead of identifying the unknown system parameters, the gains of the control law are directly adjusted by an adaptive update law without any intermediate calculation so that the desired tracking performance is achieved.

In this paper, we design a combined direct-indirect adaptive robust control method for the task of underactuated brachiation on flexible cables with parametric uncertainties and non-parametric disturbances. The proposed method relies on both estimation of the physical cable parameters and direct modification of the robust control gain. An indirect adaptive control method is developed to update the parameters of an approximate cable model, which generates a command control signal using estimates of the unknown plant dynamics, leading to a reduced tracking error.
We then design a boundary layer-based sliding mode control to robustify the system to unstructured model uncertainties and the remaining cable interaction forces, i.e., the time-varying disturbances caused by the unmodeled dynamics. The robust gain is updated by a direct adaptive control method incorporated within the feedback loop to further reduce any residual tracking error.
In the proposed unified control architecture, the adaptive control signal achieves tracking performance using parameter estimation and adaptation with no prior knowledge of the parameter variation bounds, while the robust control term guarantees tracking performance in the presence of bounded model uncertainties and independent of the parameter estimation performance.
Stability analysis of the proposed controller along with
derivation of the adaptation update laws is provided using a
Lyapunov analysis. Through simulation experiments on a full-cable model and comparison to a baseline controller, it is
shown that the proposed cable estimation-based adaptive robust controller provides reliable tracking performance and adaptive system identification for wire-borne underactuated brachiating robots, while guaranteeing
robustness to unmodeled dynamics, parametric uncertainties and unstructured disturbances.

In summary, the main contributions of this work include: i) a novel estimation-based approach to model the interactions between the flexible cable dynamics and the robot without using any sensors,
ii) formulation of a combined direct-indirect adaptive robust control scheme for wire-borne underactuated brachiating robots in the presence of parametric uncertainties and unmodeled dynamics,
iii) formal stability analysis and adaptation law derivations for the proposed control design using a Lyapunov stability argument, and iv) demonstration of the superiority of the proposed controller over the widely used input-output feedback linearization method for underactuated systems.
The proposed design leads to an adaptive robust control framework that compensates for the unknown cable force without knowing the bound of discrepancy between the approximated and actual force a priori, enabling underactuated brachiating robots to traverse flexible cables in an on-line fashion. To our knowledge, this work provides the first adaptive robust control design in the domain of underactuated brachiating robots.

\section{ROBOT-CABLE MODEL AND DYNAMICS} \label{sec:dynamics}
\subsection{Low-Fidelity Dynamic Model and Equations of Motion}
Several analytical methods such as lumped-mass model~\cite{Farzan18} and finite element method~\cite{BucFre04}
have been proposed to derive deterministic dynamics models for flexible cables. However, including a high-fidelity flexible cable dynamics into a robot model results in a large number of generalized coordinates, making it impractical for a feedback control design. Moreover, for many systems interacting with a soft body such as a cable, the configuration of the flexible body over time is of lesser importance, if any at all. Instead, the main objective of a control design is to achieve regulation or tracking for the main states of the system or robot. In such cases, it will be sufficient to only capture the effects of the flexible body on the dynamics of the system of interest.

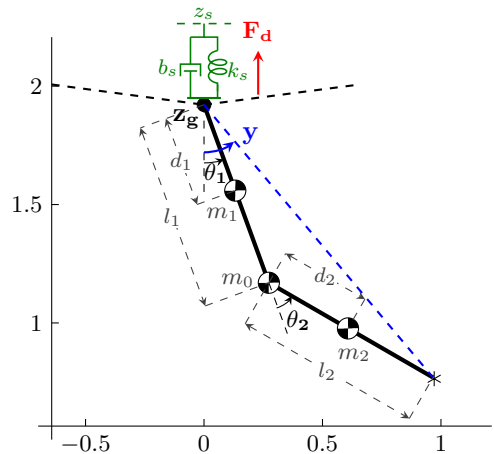
\begin{figure}[!b]
\vspace{-2pt}
\centering
\begin{tikzpicture}[scale=0.9,>=stealth]
    \tikzstyle{spring}=[semithick,decorate,decoration={aspect=0.55, segment length=1.2mm, amplitude=1.2mm,coil,mirror}]
    \tikzstyle{damper}=[semithick,decoration={markings,  
      mark connection node=dmp,
      mark=at position 0.5 with 
      {
        \node (dmp) [semithick,inner sep=0pt,transform shape,rotate=90,minimum width=7pt,minimum height=2pt,draw=none] {};
        \draw [semithick,mygreen] ($(dmp.north east)+(-2pt,0)$) -- (dmp.south east) -- (dmp.south west) -- ($(dmp.north west)+(-2pt,0)$);
        \draw [semithick,mygreen] ($(dmp.north)+(0,-2.5pt)$) -- ($(dmp.north)+(0,2.5pt)$);
      }
    }, decorate]
    \begin{scope}
    \pgfmathsetmacro{\AngleI}{-70}
    \pgfmathsetmacro{\AngleII}{-30}
    \coordinate (centro) at (2.25,4.75);
    \coordinate (centro2) at (6.75,3.5);
    \draw[ultra thick] (centro) -- ++(\AngleI:2.8) node(rod1) {} coordinate (joint1);
    \draw[ultra thick] (joint1) -- ++(\AngleII:2.8) node(rod2) {} coordinate (joint2);
    \filldraw [fill=black,draw=black] (centro) circle[radius=0.1];
    \pgfsetplotmarksize{0.7ex}
    \pgfplothandlermark{\pgfuseplotmark{asterisk}}
    \pgfplotstreamstart
    \pgfplotstreampoint{\pgfpoint{5.65cm}{0.7cm}}
    \pgfplotstreamend
    \path (centro) +(\AngleI:1.35) coordinate (com1);
    \path (joint1) +(\AngleII:1.35) coordinate (com2);
    \node [] at (joint1) {\centerofmass};
    \node [] at (com1) {\centerofmass};
    \node [] at (com2) {\centerofmass};
    \node at (joint1) [xshift=-2pt,yshift=1pt] [left,darkgray] {\small $m_{0}$};
    \node at (com1) [xshift=5pt,yshift=-3pt] [below left,darkgray] {\small $m_{1}$};
    \node at (com2) [xshift=12pt,yshift=-4pt] [below left,darkgray] {\small $m_{2}$};
    \node at (centro) [xshift=2pt,yshift=2pt] [below left] {$\mathbf{z_g}$};
    \draw (-0.2,0) -- (6.5,0);
    \draw (0,-0.2) -- (0,5.75);
    \draw [thick, dashed] (0,5.05) -- (centro);
    \draw [thick, dashed] (centro) -- (4.55,5.05);
    \draw[dashed,-] (centro) -- ++ (0,-1.4) node (mary1) [below]{};
    \draw[dashed,-] (joint1) -- +(\AngleI:0.8) node (mary2) [below]{};
    \path (centro) +(90+\AngleI:-0.6) coordinate (d11);
    \path (com1) +(90+\AngleI:-0.6) coordinate (d12);
    \path (joint1) +(90+\AngleII:0.5) coordinate (d21);
    \path (com2) +(90+\AngleII:0.5) coordinate (d22);
    \path (centro) +(90+\AngleI:-1) coordinate (l11);
    \path (joint1) +(90+\AngleI:-1) coordinate (l12);
    \path (joint1) +(90+\AngleII:-0.7) coordinate (l21);
    \path (joint2) +(90+\AngleII:-0.7) coordinate (l22);
    \draw[<->,dashed,darkgray] (d11) -- (d12) node [midway,fill=white,inner sep=1pt]{\footnotesize $d_1$};
    \draw[<->,dashed,darkgray] (d21) -- (d22) node [midway,fill=white,inner sep=0pt]{\footnotesize $d_2$};
    \draw[<->,dashed,darkgray] (l11) -- (l12) node [midway,fill=white,inner sep=1pt]{\footnotesize $l_1$};
    \draw[<->,dashed,darkgray] (l21) -- (l22) node [midway,fill=white,inner sep=1pt]{\footnotesize $l_2$};
    \draw[dashed,darkgray] (centro) -- (l11);
    \draw[dashed,darkgray] (d12) -- (com1);
    \draw[dashed,darkgray] (l12) -- (joint1);
    \draw[dashed,darkgray] (l21) -- (d21);
    \draw[dashed,darkgray] (d22) -- (com2);
    \draw[dashed,darkgray] (l22) -- (joint2);
    \draw[dashed, thick, blue] (centro) -- (joint2) node (mary3) [below]{};
    \node at (centro) [xshift=11pt,yshift=-5pt,blue] [below right] {$\mathbf{y}$};
    \pic [draw, ->, thick, angle eccentricity=1.25, angle radius=18, blue] {angle = mary1--centro--mary3};
    \pic [draw, ->, "\small $\mathbf{\theta_1}$", angle eccentricity=1.2, angle radius=22] {angle = mary1--centro--rod1};
    \pic [draw, ->, "\small $\mathbf{\theta_2}$", angle eccentricity=1.8, angle radius=10] {angle = mary2--joint1--rod2};
    \foreach \x/\xtext in {0.5/-0.5, 2.25/0, 4/0.5, 5.75/1}
      \draw[shift={(\x,0)}] (0pt,3pt) -- (0pt,0pt) node[below] {\small $\xtext$};
    \foreach \y/\ytext in {1.525/1 , 3.275/1.5, 5.025/2}
      \draw[shift={(0,\y)}] (3pt,0pt) -- (0pt,0pt) node[left] {\small $\ytext$};
   \end{scope}
   \begin{scope}
    \draw [semithick,mygreen] (centro)++(0,1.2) -- ++(0,-0.2) -- ++(0.2,0) -- ++(0,-0.2) coordinate (p1);
    \draw [spring,mygreen] (p1) -- ++(0,-0.6) node [midway,right,mygreen]{\footnotesize $k_s$};
    \draw [semithick,mygreen] (centro)++(0,0.1) -- ++(0.2,0) -- ++(0,0.2);
    \draw [semithick,mygreen] (centro)++(0,0.1) -- ++(-0.2,0) coordinate (d1) ++(0,0.9) coordinate (d2);
    \draw [damper,mygreen] (d1) -- ($(d1)!(d2)!(d1)$) node [midway,left,mygreen,xshift=-1pt]{\footnotesize $b_s$};
    \draw [semithick,mygreen] (d2)  -- ++(0.2,0);
    \draw [semithick,dashed,mygreen] (centro)++(-0.4,1.2)  -- ++(0.8,0) node [midway,above,mygreen,yshift=-2pt]{\footnotesize $z_{s}$};
    \draw [thick,mygreen] (centro)++(-0.25,0.1)  -- ++(0.5,0);
    \draw[thick, -stealth, red] (centro)+(0.8,0.15) -- ++ (0.8,0.8) node (force) [above]{\small $\mathbf{F_d}$};
   \end{scope}
\end{tikzpicture}
\vspace{-4pt}
\caption{The 3-DOF model of the two-link brachiating robot attached to a flexible cable. The flexible cable is modeled as a parallel spring-damper with uncertain parameters and a residual force applied to the pivot gripper.}
\label{fig:robot-schematic}
\end{figure}

Attaching a brachiating robot to a flexible cable results in two major differences when compared to a robot attached to a rigid bar: an additional generalized coordinate as the cable gives the pivot gripper (grasping the cable) the freedom to go up and down, and a residual force applied to the pivot gripper by the cable. Fig. \ref{fig:robot-schematic} presents our proposed low-fidelity model for a wire-borne underactuated brachiating robot. Using this model, the dynamic effects of the flexible cable are captured as a time-varying force $F_c$ applied to the pivot gripper~\cite{Farzan20}:
\begin{equation}
    F_c(t) = F_s(t) + F_d(t), \label{eq:cable-force}
\end{equation}
where $F_s$ is the force generated by a spring-damper -- with one end attached to the pivot gripper and the other end fixed at a specific height, and $F_d$ is the residual force applied by the cable to the pivot gripper.

The physical parameters of the spring-damper model
include $k_s$ as the stiffness of the spring, $b_s$ as the damping value of the damper, and $z_s$ as the attachment height of the fixed end of the spring. Hence, the spring-damper force $F_s$ applied to the robot can be written as:
\begin{equation}
    F_s = k_s(z_s-z_g) + b_s(-\dot{z}_g). \label{eq:spring-force}
\end{equation}

While a single spring-damper provides the additional generalized coordinate introduced by the cable, it only approximates the dominant harmonic frequency of the cable, and hence the residual force $F_d$ is added as an external disturbance to account for all other harmonics and dynamic effects.
The proposed model provides the ability to include parametric model uncertainties as well as unstructured (non-parametric) disturbances in the state equations,
paving the way for an \emph{adaptive-robust} feedback control design.

The model shown in Fig. \ref{fig:robot-schematic} consists of 3 degrees-of-freedom (DOFs): $\theta_1$ as the angle between the pivot arm and a vertical line, $\theta_2$ as the angle between the swing arm and the pivot arm, and $z_g$ as the vertical Cartesian position of the pivot gripper. Therefore, the state vector of the system is represented by $x=[\theta_1,\,\theta_2,\,z_g,\,\dot{\theta}_1,\,\dot{\theta}_2,\,\dot{z}_g]^T$. There is a single actuator exerting a torque about the center joint, hence the system possess 2 degrees of underactuation.

The nonlinear equations of motion for the robot-cable system are derived via the Lagrangian method and written in the general manipulator form of $M(q)\ddot{q}+C(q,\dot{q})\dot{q}+D(q){=}\tau$:

\noindent \small \begin{gather}
    \hspace{-6pt} m_{11}(q)\ddot{q}_1+m_{12}(q)\ddot{q}_2+m_{13}(q)\ddot{q}_3+c_1(q,\dot{q})\dot{q}+d_1(q)=0 \label{eq:eom1}, \\
    \hspace{-6pt} m_{21}(q)\ddot{q}_1+m_{22}(q)\ddot{q}_2+m_{23}(q)\ddot{q}_3+c_2(q,\dot{q})\dot{q}+d_2(q)=u \label{eq:eom2}, \\
    \hspace{-6pt} m_{31}(q)\ddot{q}_1+m_{32}(q)\ddot{q}_2+m_{33}(q)\ddot{q}_3+c_3(q,\dot{q})\dot{q}+d_3(q)=F_c \label{eq:eom3}, 
\end{gather} \normalsize

\noindent where $q{:=}[q_1,\, q_2,\, q_3]^T{=}[\theta_1,\, \theta_2,\, z_g ]^T$ represents the vector of generalized coordinates, $u\in\mathbb{R}$ is the input torque, and $F_c\in\mathbb{R}$ is the force in (\ref{eq:cable-force}) applied by the cable to the pivot gripper. $M(q)\in\mathbb{R}^{3\times 3}$, $C(q,\dot{q})\in\mathbb{R}^{3}$ and  $D(q)\in\mathbb{R}^{3}$ represent the positive definite inertial matrix, the Coriolis and centripetal forces, and the gravitational forces, respectively.

\subsection{Output Definition and Output Dynamics} \label{subsec:output}
A successful brachiating swing can be described as starting from an initial configuration $x_0$ where both grippers are attached to the cable and then releasing the swing gripper and applying a control input for a short finite time horizon such that the swing gripper reaches and grabs the cable on the other side of the pivot gripper. For such motions going from left to right, we note that the angle between the vertical axis and the line connecting the pivot gripper to the swing gripper (shown with the blue dashed line in Fig. \ref{fig:robot-schematic}) always starts about $-90$ degrees and ends about $90$ degrees. We define this angle as the output of the system and denote it by $y$. For a robot with equal arm lengths ($l_1=l_2$ in Fig. \ref{fig:robot-schematic}), the output angle $y\in\mathbb{R}$ can be defined in terms of the generalized coordinates of the system as:
\begin{equation} 
    y=f_{un}(\theta_1,\,\theta_2)=q_1+0.5\,q_2. \label{eq:ydef} 
\end{equation}

To obtain the output dynamics $\ddot{y}$, we solve (\ref{eq:eom3}) for $\ddot{q}_3$ and substitute the resulting expression into (\ref{eq:eom1}) and (\ref{eq:eom2}) to eliminate the  $\ddot{q}_3$ variable.
Having two equations
and two unknowns ($\ddot{q}_1$ and $\ddot{q}_2$), we solve for $\ddot{q}_1$ and $\ddot{q}_2$ and substitute the results into the equation $\ddot{y}=\ddot{q}_1+\frac{1}{2}\ddot{q}_2$ to get the nonlinear output dynamics as a function of the system state vector $x$:
\begin{equation}
    \ddot{y}=f\big(x,\,F_c(t),\,u\big). \label{eq:ydd}
\end{equation}

\begin{remark}
For the current work, we have made the assumption that all the system states are measurable.  In general, for applications with unmeasurable states, estimation methods like Extended Kalman Filter (EKF)~\cite{Simon06} can be used to estimate the unmeasurable states.
\end{remark}

\section{PROPOSED CONTROLLER AND STABILITY ANALYSIS}
This section presents the combined direct-indirect adaptive robust control design and provides stability analysis and adaptation laws derivation for the proposed controller.
We first develop an input-output linearization of the system's output to derive the linear error dynamics. We then present the adaptive estimation and the robust control design to estimate the parameter uncertainties of the cable and compensate for any non-parametric disturbances introduced by the residual force $F_d$. Lastly, the stability analysis and derivation of the adaptation update laws will be presented.

\subsection{Control Law and Error Dynamics}
Considering the output $y$ for the robot-cable system derived in Section \ref{sec:dynamics},
the nonlinear output dynamics in (\ref{eq:ydd}) can be linearly parameterized into a control-affine form as:
\begin{equation}
    \ddot{y}(t) = g(\theta) + h(x)p + \beta(\theta) F_d(t) + \alpha(\theta) u, \label{eq:ydd0}
\end{equation}
where the functions $g(\cdot)$, as well as $h(\cdot)$, $\alpha(\cdot)$ and $\beta(\cdot)$ coefficients are known nonlinear functions of either the state vector $x$ or the sub-state $\theta=[\theta_1,\,\theta_2,\,\dot{\theta}_1,\,\dot{\theta}_2]^T$.
The vector $p\in\mathbb{R}^3$ contains the unknown spring-damper model parameters:\looseness=-1
\begin{equation}
    p = \begin{bmatrix} k_{s} & b_{s} & k_{s}z_{s} \end{bmatrix}^T.
\end{equation}
As shown in (\ref{eq:spring-force}), the parameter $z_s$ is always multiplied by $k_s$, hence we consider the term $k_sz_s$ as a separate parameter.

Given a desired trajectory for $y$, denoted by $y_d$, and dropping the time-varying notation, we define the tracking error $e$, the error velocity $\dot{e}$, and the error dynamics $\ddot{e}$ as:
\begin{gather}
    e = y_d - y, \qquad \dot{e}=\dot{y}_d - \dot{y}, \\
    \ddot{e} = \ddot{y}_d - \ddot{y} = \ddot{y}_d - g(\theta) - h(x)p - \beta(\theta) F_d - \alpha(\theta) u. \label{eq:error-dynamics3}
\end{gather}
Therefore, the control law:
\begin{equation}
    u = \alpha^{-1}(\theta)\big(\ddot{y}_d-g(\theta)- h({x})\hat{p}+v +\lambda\dot{e} \big), \label{eq:control-law3}
\end{equation}
input-output linearizes the system, with
the vector $\hat{p}\in\mathbb{R}^3$ as the estimate of the uncertain parameters $p$. The control law (\ref{eq:control-law3}) includes two parts: i) the term $h({x})\hat{p}$ represents the indirect adaptive component, which will be estimated by an adaptation scheme to compensate for the uncertain cable parameters; ii) the second part, $v + \lambda \dot{e}$ is the robust control term which will be generated by a boundary layer-based sliding mode scheme with direct adaptation to robustify the closed-loop system against the adaptive estimation errors and the variations of the residual force $F_d$ as a non-parametric disturbance.\looseness=-1

Substituting (\ref{eq:control-law3}) into (\ref{eq:error-dynamics3}) results in the error dynamics:
\begin{equation}
    \ddot{e} = -h(x)\tilde{p} - \beta(\theta) F_d - v -\lambda\dot{e},
\end{equation}
where the vector $\tilde{p} = p - \hat{p}$ represents the parameter estimation error in the system as the parametric uncertainties.
Perfect desired trajectory tracking and reaching the desired final configuration are achieved by $[e, \, \dot{e}]^T \to 0$.

\subsection{Direct-Indirect Adaptive Robust Control Design}\label{subsec:controller}
Defining the time-varying error variable $s\in\mathbb{R}$ as the weighted sum of the position error $e$ and the velocity error $\dot{e}$:\looseness=-1

\noindent \begin{gather}
    s = \dot{e} + \lambda e, \quad \lambda>0, \label{eq:seq}
\end{gather}
the problem of tracking the two-dimensional vector $[e, \, \dot{e}]^T$ is reduced to that of keeping the scalar quantity $s$ at zero.

We define the robust control variable $v$ as:
\begin{equation}
    v = k_d(t)\,sat(\nicefrac{s}{\phi}), \label{eq:vrbst}
\end{equation}
in which $k_d(t)$ is a direct adaptive gain, compensating for unmodeled dynamics and the disturbance $F_d$. The function $sat(\cdot)$ represents the saturation function, with $\phi\in\mathbb{R}^+$ as the width of saturation:
\begin{equation}
    sat(\nicefrac{s}{\phi})=\begin{cases}
    \nicefrac{s}{\phi} & \textrm{if} \; \abs{\nicefrac{s}{\phi}}\leq1 \\
    sgn(\nicefrac{s}{\phi}) & \textrm{otherwise} \\
    \end{cases}.
\end{equation}

\noindent Substituting the control term in (\ref{eq:vrbst}) into (\ref{eq:error-dynamics3}), we obtain the error dynamics as:
\begin{equation}
    \ddot{e} = -h(x)\tilde{p} - \beta(\theta) F_d - k_d\,sat(\nicefrac{s}{\phi}) -\lambda\dot{e},
\end{equation}
and the error trajectories $[e, \, \dot{e}]^T$ can be expressed directly in terms of the variable $s$ as:
\begin{equation}
    \dot{s} = \ddot{e} + \lambda \dot{e} = -h({x})\tilde{p} - \beta(\theta) F_d - k_d\,sat(\nicefrac{s}{\phi}). \label{eq:sdot}
\end{equation}

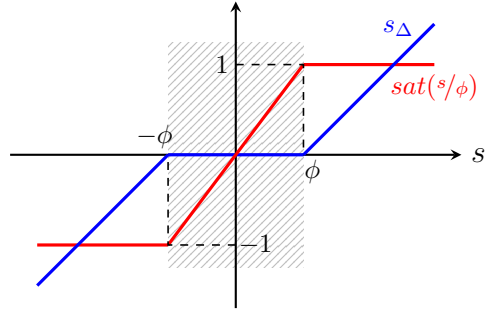
\begin{figure}[t]
    \centering
    \begin{tikzpicture}[scale=1.2]
    \def\phiv{0.75}
    \pattern[pattern=north east lines, pattern color=lightgray] (-\phiv,-1.25)--(-\phiv,1.25)--(\phiv,1.25)--(\phiv,-1.25)--cycle;
    \draw[-stealth,thick] (-2.5,0)--(2.5,0) node[right]{\large $s$};
    \draw[-stealth,thick] (0,-1.7)--(0,1.7) node[above]{};
    \draw[domain=-\phiv:\phiv,smooth,variable=\x,blue, very thick] plot ({\x},{0});
    \draw[domain=-\phiv:\phiv,smooth,variable=\x,red, very thick] plot ({\x},{\x/\phiv});
    \draw[domain=\phiv:2.2,smooth,variable=\x,red, very thick] plot ({\x},{1});
    \draw[domain=-\phiv:-2.2,smooth,variable=\x,red, very thick] plot ({\x},{-1});
    \draw[domain=\phiv:2.2,smooth,variable=\x,blue, very thick] plot ({\x},{\x-\phiv});
    \draw[domain=-\phiv:-2.2,smooth,variable=\x,blue, very thick] plot ({\x},{\x+\phiv});
    \draw[dashed, semithick] (0,1)--(\phiv,1)--(\phiv,0);
    \draw[dashed, semithick] (0,-1)--(-\phiv,-1)--(-\phiv,0);
    \node at (\phiv+0.1,-0.18) {$\phi$};
    \node at (-\phiv-0.15,0.2) {$-\phi$};
    \node at (-0.15,1) {$1$};
    \node at (0.22,-1) {$-1$};
    \node at (1.8,1.4) [blue] {$s_{\Delta}$};
    \node at (2.2,\phiv) [red] {$sat(\nicefrac{s}{\phi})$};
    \end{tikzpicture}
    \caption{Saturation function $sat(\cdot)$ and the boundary layer trajectory $s_{\Delta}$. The boundary layer $s \leq \abs{\phi}$ is represented by the hatched area.}
    \label{fig:sDelta}
\end{figure}

Adaptation update laws for the estimated vector $\hat{p}$ and for the gains $k_d$ can be derived by selecting a proper Lyapunov function. As will be shown in Section \ref{subsec:stability},
to achieve the best approximation of $p$ and $k_d$, the following adaptation laws are suggested for the system:
\begin{gather}
    \dot{\hat{p}} = -\Gamma {h({{x}})}^T s_{\Delta}, \label{eq:update-law1}\\
    \dot{k}_d = k_{d_0} \abs{s_{\Delta}}, \label{eq:update-law2}
\end{gather}
where $\Gamma\in\mathbb{R}^{3\times 3}$ is a positive definite matrix with diagonal elements containing the indirect adaptive gains adjusted by user, $k_{d_0}$ is the initial guess for the robust control gain, and $s_{\Delta}$ is the \emph{boundary layer trajectory}~\cite{SloCoe86}, defined as:
\begin{equation}
    s_\Delta =
    \left\{
    \begin{array}{ll}
    0, & \abs{s}\leq \phi \\
    s- \phi\,sat(\nicefrac{s}{\phi}) & \abs{s}>\phi
    \end{array}
    \right. . \label{eq:sDelta}
\end{equation}
We define the region $\abs{s}\leq \phi$ as the \emph{boundary layer} in order to stop the direct and indirect adaptation mechanisms inside the boundary layer and impose a trade-off between tracking performance and control chattering. An example of $s_{\Delta}$ is illustrated in Fig. \ref{fig:sDelta}.

The direct and indirect adaptation laws tune $k_d(t)$ and $\hat{p}$ respectively for sufficiently large tracking error ($\abs{s}>\phi$), and stop tuning when the error trajectory lies inside the boundary layer ($-\phi\leq s \leq \phi$). Small tracking errors are usually caused by the measurement noise, to which the adaptation mechanisms should be insensitive.

The proposed controller shows robustness to the parametric uncertainties $\tilde{p}$ and non-parametric disturbance $F_d$, while all the error trajectories starting outside the boundary layer converge to the boundary layer and those starting inside the boundary layer remain inside for all $t \geq 0$. To prove stability and tracking performance of the proposed controller, the following scalar positive definite continuously-differentiable Lyapunov function is considered:
\begin{equation}
    V(s_{\Delta},\tilde{p},\tilde{k}_d) = \frac{1}{2} s_{\Delta}^2 + \frac{1}{2} \tilde{p}^T\Gamma^{-1}\tilde{p}
    + \frac{1}{2}\tilde{k}_d^2, \label{eq:lyapunov}
\end{equation}
with $\tilde{k}_d(t) = k_d(t) - \bar{k}_d$ where $\bar{k}_d$ is the unknown positive ideal robust gain and $\tilde{k}_d(t)$ is the gain estimation error.

\begin{theorem}
Given the Lyapunov function in (\ref{eq:lyapunov}), the control law (\ref{eq:control-law3}) when used in conjunction with the update laws (\ref{eq:update-law1}) and (\ref{eq:update-law2}), and applied to the closed-loop system (\ref{eq:sdot}), results in $\dot{V}(s_{\Delta},\tilde{p},\tilde{k}_d)\to 0$ as $t\to \infty$, which implies that $s_{\Delta}\to 0$, i.e. $\abs{s} \leq \phi$ and $\abs{e} \leq \frac{\phi}{\lambda}$, so that the closed-loop system is asymptotically stable,
and both the parameter estimation error vector $\tilde{p}$ and the gain estimation error $\tilde{k}_d$ are bounded.
\end{theorem}

For the proof of Theorem 1, see Section \ref{subsec:stability}.

\subsection{Stability Analysis and Adaptation Laws Derivation} \label{subsec:stability}
We now present the proof of Theorem 1, which provides the stability analysis of the adaptive-robust control law proposed in Section \ref{subsec:controller}, and consequently the derivation of the direct-indirect adaptation update laws for estimation of the cable force and robustness to unmodeled dynamics and residual disturbances.

\begin{proof}[Proof of Theorem 1]
Although $s_{\Delta}$ is not differentiable everywhere, the Lyapunov candidate $V(s_{\Delta},\tilde{p},\tilde{k}_d)$ in (\ref{eq:lyapunov}) is differentiable as it is a quadratic function of $s_{\Delta}$. The derivative of the Lyapunov function $V$ for the active boundary region (i.e. outside of the boundary layer) is computed as follows:\looseness=-1
\begin{equation}
    \dot{V}(s_{\Delta},\tilde{p},\tilde{k}_d) = s_{\Delta}\dot{s}_{\Delta} -
    \tilde{p}^T\Gamma^{-1}\dot{\hat{p}}+\tilde{k}_d\dot{k}_d.
\end{equation}

\subsubsection*{Case 1} For the \textit{outside} of the boundary layer ($\abs{s}>\phi$), from (\ref{eq:sDelta}) we have $\dot{s}_{\Delta}=\dot{s}$, so using the closed-loop form in (\ref{eq:sdot}) we obtain:

\noindent \begin{align}
    \dot{V}(s_{\Delta},\tilde{p},\tilde{k}_d)
    = \; & - s_{\Delta}h({x})\tilde{p} - s_{\Delta}\beta(\theta) F_d \nonumber \\
    &  - s_{\Delta}k_d\,sat(\nicefrac{s}{\phi}) - \tilde{p}^T\Gamma^{-1}\dot{\hat{p}} + \tilde{k}_d\dot{k}_d. \label{eq:Vdot}
\end{align}

To derive the adaptation laws, we constrain $-\tilde{p}^T\Gamma^{-1}\dot{\hat{p}} -s_{\Delta}h({x})\tilde{p}\,$ to zero. Noting that $s_{\Delta}\,sat{(\nicefrac{s}{\phi})} = |s_{\Delta}|$, the adaptation laws are formulated as:
\begin{gather}
    \dot{\hat{p}}=-\Gamma h({x})^Ts_{\Delta}, \\
    \dot{k}_d = k_{d_0}\abs{s_{\Delta}},
\end{gather}
as already presented in (\ref{eq:update-law1}) and (\ref{eq:update-law2}).

\noindent Substituting the adaptation update laws into (\ref{eq:Vdot}) gives:
\begin{align}
    \hspace{-0.1in} \dot{V}(s_{\Delta},\tilde{p},\tilde{k}_d) 
    & = - s_{\Delta}\beta(\theta) F_d - k_d\abs{s_{\Delta}} + k_{d_0}\tilde{k}_d \abs{s_{\Delta}} \nonumber \\
    & \hspace{-25pt} \leq
    \abs{s_{\Delta}}\abs{\beta(\theta) F_d} - k_d\abs{s_{\Delta}} + k_{d_0}(k_d - \bar{k}_d) \abs{s_{\Delta}}.
\end{align}
Defining the matching condition as:
\begin{equation}
    \bar{k}_d = \frac{1}{k_{d_0}} \abs{\beta(\theta) F_d},
\end{equation}
results in:
\begin{equation}
    \dot{V}(s_{\Delta},\tilde{p},\tilde{k}_d) \leq -k_d\abs{s_{\Delta}} +  k_{d_0}k_d\abs{s_{\Delta}} = -k_d\abs{s_{\Delta}}\big(1 - k_{d_0}\big).
\end{equation}

\noindent Therefore, by choosing the tuning parameter $k_{d_0} < 1$, it is guaranteed that:
\begin{equation}
    \dot{V}(s_{\Delta},\tilde{p},\tilde{k}_d) \leq -k_d\abs{s_{\Delta}}\big(1 - k_{d_0}\big) \leq 0.
\end{equation}
This indicates that outside the boundary layer, the Lyapunov derivative is negative semi-definite.

\begin{lemma}[Barbalat's Lemma]
    If a function $\phi(t)$ is uniformly continuous\footnote{A function $\phi(t): \mathbb{R}\to\mathbb{R}$ is uniformly continuous on $[0,\,\infty]$ if $\forall \epsilon>0, \; \exists \, \delta(\epsilon)>0$ such that $\forall t_1\geq 0,\,\forall t_2\geq 0$, $\abs{t_1-t_2} \leq \delta$, then $\abs{\phi(t_1)-\phi(t_2)}\leq\epsilon$. Boundedness of the derivative $\dot{\phi}(t)$ implies uniform continuity.}
    for all $t \geq 0$, and if the limit of the integral $\lim_{t\to\infty}\int_0^t \phi(\tau)d\tau$
    exists and is finite, then $\lim_{t\to\infty} \phi(t)=0$.
\end{lemma}

Let us define:
\begin{equation}
    \phi(t) = k_d\abs{s_{\Delta}}\big(1 - k_{d_0}\big),
\end{equation}
based on which it follows that:
\begin{equation}
    \dot{V}(s_{\Delta},\tilde{p},\tilde{k}_d) \leq -\phi(t). \label{eq:Vdotphi}
\end{equation}
Integrating both sides of (\ref{eq:Vdotphi}) from $0$ to $\infty$ yields:
\begin{equation}
    V(0) - V(\infty) \geq \lim_{t\to\infty}\int_0^t \phi(\tau)d\tau. \label{eq:Vint}
\end{equation}

Since $\dot{V} \leq 0$ and by definition $V \geq 0$, the left-hand side of (\ref{eq:Vint}) exists, and is positive and finite. Hence, according to Barbalat's Lemma:
\begin{equation}
    \lim_{t\to\infty} \phi(t) = \lim_{t\to\infty} \Big( k_d\abs{s_{\Delta}}\big(1 - k_{d_0}\big) \Big)=0. \label{eq:limV}
\end{equation}
Since $k_d>0$ and $k_{d_0}<1$, (\ref{eq:limV}) implies that $s_{\Delta} \to 0$ asymptotically, which follows that $\abs{s} \leq \phi$ and $\abs{e} \leq \frac{\phi}{\lambda}$ according to (\ref{eq:sDelta}) and (\ref{eq:seq}) respectively.

Similarly, negative semi-definiteness of $\dot{V}$ and positive definiteness of $V$ imply that $0 < V(t) \leq V(0) \;\, \forall t\geq 0$, which follows that $V$ is upper and lower bounded, i.e. bounded. Boundedness of $V$ implies that $s_{\Delta}$, $\tilde{p}$ and $\tilde{k}_d$ are bounded. As we  already showed that $s_{\Delta}$ converges to zero, we conclude that $\tilde{p}$ and $\tilde{k}_d$ are bounded.

This implies that under the proposed adaptive-robust controller, by choosing $k_{d_0}<1$, the system's solution $s_{\Delta}$ is convergent, and $\tilde{p}$ and $\tilde{k}_d$ are bounded.

\subsubsection*{Case 2} For the \textit{inside} of the boundary layer ($\abs{s} \leq \phi$), from (\ref{eq:sDelta}) we have ${s}_{\Delta}=0$, so substituting the adaptation update laws (\ref{eq:update-law1}) and (\ref{eq:update-law2}) in (\ref{eq:lyapunov}) results in:
\begin{equation}
    \dot{V}(s_{\Delta},\tilde{p},\tilde{k}_d)=0.
\end{equation}
This implies that the boundary layer is an invariant set, and all the error trajectories starting inside the boundary layer remain inside for all $t \geq 0$.

The combination of \textit{Case 1} and \textit{Case 2} proves the stability of the proposed adaptive-robust controller, regardless of starting point. By selecting the initial adaptive gain such that $k_{d_0}<1$, the error variable $s$ and the error trajectories are convergent (i.e. $\abs{s} \leq \phi$ and $\abs{e} \leq \frac{\phi}{\lambda}$), while the parameter estimation error vector $\tilde{p}$ and the gain estimation error $\tilde{k}_d$ are bounded.
\end{proof}

\section{RESULTS AND DISCUSSION}
In this section, we evaluate the proposed direct-indirect adaptive robust control through simulation experiments for brachiation on an 8 meter flexible cable with unknown dynamics. 
For the simulation settings, the dynamic model of the flexible cable is derived using the finite element method~\cite{Farzan18}.
However, for all the experiments presented in this section, the flexible cable model and its equivalent spring-damper parameters are unknown to the controller, and the control design is based on the spring-damper-force model with unknown parameters presented in section \ref{sec:dynamics}.
The physical parameters of the robot-cable system used in the simulations are listed in Table \ref{tab:physical-parameters}.\looseness=-1

\begin{table}[!b]
\caption{\label{tab:physical-parameters}\vspace{-3pt}Physical parameters of the robot and the cable}
\noindent \begin{centering}
\begin{tabular}{|c|c|}
\hline 
{\textit{\footnotesize{}Parameter}} & \textit{\footnotesize{}Value}\tabularnewline
\hline 
{\footnotesize{}Main body center of mass} & {\footnotesize{}$m_{0}=2\,\textrm{kg}$}\tabularnewline
\hline 
{\footnotesize{}Link 1 and 2 center of mass} & {\footnotesize{}$m=m_{1}=m_{2}=1\,\textrm{kg}$}\tabularnewline
\hline 
{\footnotesize{}Link 1 and 2 length} & {\footnotesize{}$l=l_{1}=l_{2}=0.71\,\textrm{m}$}\tabularnewline
\hline 
{\footnotesize{}Link 1 center of mass location} & {\footnotesize{}$d_{1}=\nicefrac{2}{3}\,l_{1}$} \tabularnewline
\hline 
{\footnotesize{}Link 2 center of mass location} & {\footnotesize{}$d_{2}=\nicefrac{1}{3}\,l_{2}$} \tabularnewline
\hline 
{\footnotesize{}Link 1 and 2 moment of inertia } & {\footnotesize{}$I_{1}=I_{2}=\nicefrac{1}{12}\,m l^{2}$}\tabularnewline
\hline 
{\footnotesize{}Cable length} & {\footnotesize{}$l_{c}=8\,\textrm{m}$} \tabularnewline
\hline 
{\footnotesize{}Cable linear mass} & {\footnotesize{}$m_{c}=0.25\,\textrm{kg/m}$} \tabularnewline
\hline 
{\footnotesize{}Cable stiffness} & {\footnotesize{}$k_{c}=785400\,\textrm{N/m}$} \tabularnewline
\hline 
{\footnotesize{}Cable damping} & {\footnotesize{}$b_{c}=4\,\textrm{N\,s/m}$}\tabularnewline
\hline 
\end{tabular}
\par\end{centering}{\small \par}
\end{table}

We generated an optimal reference trajectory using a nonlinear programming trajectory optimization method~\cite{Farzan18} for the robot attached to the spring-damper model with $k_s{=}680$ N/m, $b_s{=}20$ Ns/m and $z_s{=}1.9$ m, with no residual force, performing a single brachiating maneuver for the initial and final states of $[-48^{\circ},-98^{\circ},1.84\,\textrm{m},0,0,0]$ and $[46^{\circ},96^{\circ},1.84\,\textrm{m},0,0,0]$ respectively, associated to $[\theta_1,\,\theta_2,\,z_g,\,\dot{\theta}_1,\,\dot{\theta}_2,\,\dot{z}_g]$.
We then computed a desired trajectory for the outputs $y$ and $\dot{y}$ using (\ref{eq:ydef}) based on the reference trajectory, with the finite time horizon set to $t\in[0,\,1.1]$ seconds according to the reference trajectory. The control input is constrained to $u\in[-10,10]$ N\,m, accounting for the torque limits of the actuator installed on the hardware robot.

For the experiments to follow, the initial values of the spring-damper parameters are set to $k_s=400$ N/m, $b_s=12$ N\,s/m and $z_s=1.6$ m, resulting in about ${40\%}$ uncertainty compared to the true spring-damper model.
We also tune the design parameters of the proposed controller to achieve the best balance between tracking accuracy and control chattering while keeping the control torque as optimal as possible:
error variable scaling factor $\lambda=8$,
indirect adaptive estimation gains $\Gamma=diag[100,\,10,\,100]$, boundary layer width of saturation $\phi=0.4$, and the initial robust gain $k_{d_0}=0.5$.

To provide a baseline for the performance of the proposed controller and demonstrate the importance of using an adaptive robust control design for this application, we make comparison with an input-output feedback linearization controller~\cite{SloWei91}, with PD gains set to $k_p = 20$ and $k_d = 5$. The gains are carefully tuned so that the baseline controller achieves reliable tracking when brachiating on a rigid bar, and obtains its best possible performance on a flexible cable.\looseness=-1

Simulation videos of the results presented in this section can be found
in the accompanying video for the paper, available at {\small \url{https://vimeo.com/sfarzan/cdc20}}.

\subsection{Performance Comparison with a Baseline Controller}
\begin{figure}[t]
\renewcommand{\arraystretch}{0.2}
\setlength\tabcolsep{0.5pt}
\begin{tabular}{cc}
{\begin{tikzpicture}\node[inner sep=0pt] at (0,0)
{\includegraphics[trim={5bp 1bp 50bp 20bp}, clip, width=0.5\columnwidth] {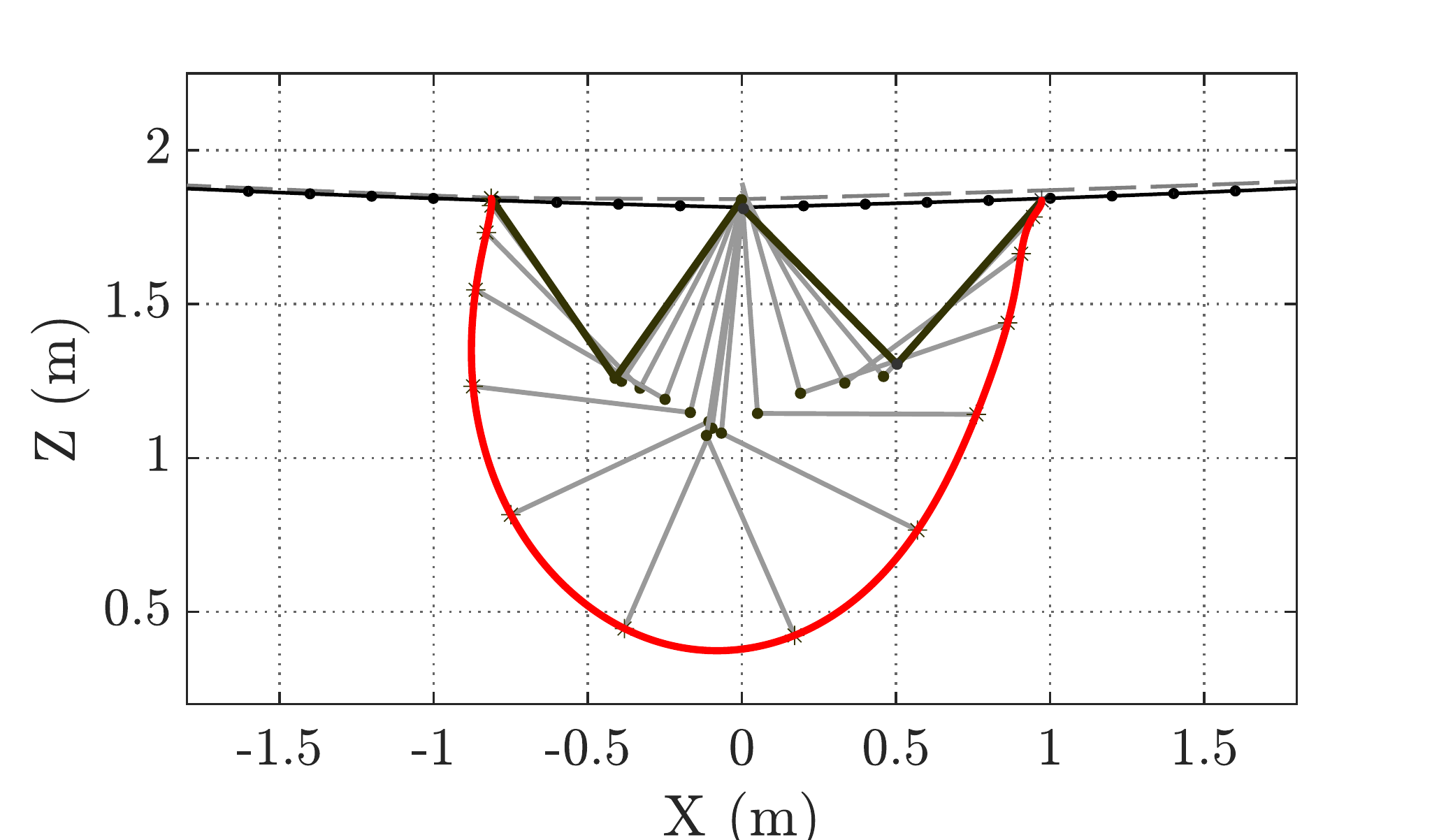}};
\draw[<-,>=stealth',semithick,red,dashed] (-0.75,-0.1) to [out=130,in=230] (-0.75,0.75);
\end{tikzpicture}} & 
{\begin{tikzpicture}\node[inner sep=0pt] at (0,0)
{\includegraphics[trim={5bp 1bp 50bp 20bp}, clip, width=0.5\columnwidth] {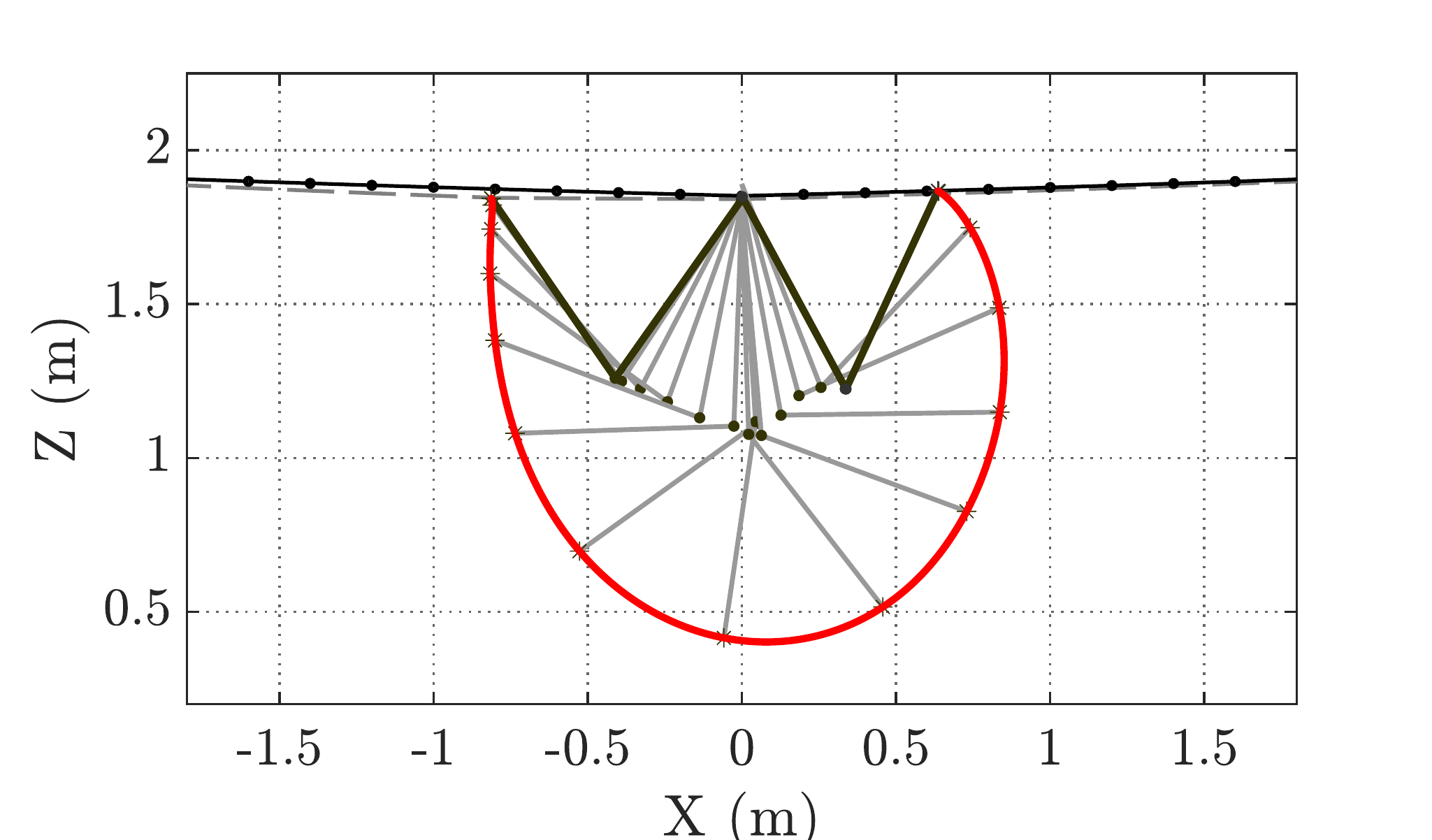}};
\draw[<-,>=stealth',semithick,red,dashed] (-0.75,-0.1) to [out=130,in=230] (-0.75,0.75);
\end{tikzpicture}}
\tabularnewline
{\scriptsize{}(a)} & {\scriptsize{}(b)}
\tabularnewline
{\includegraphics[trim={1bp 105bp 50bp 110bp},clip,width=0.5\columnwidth]{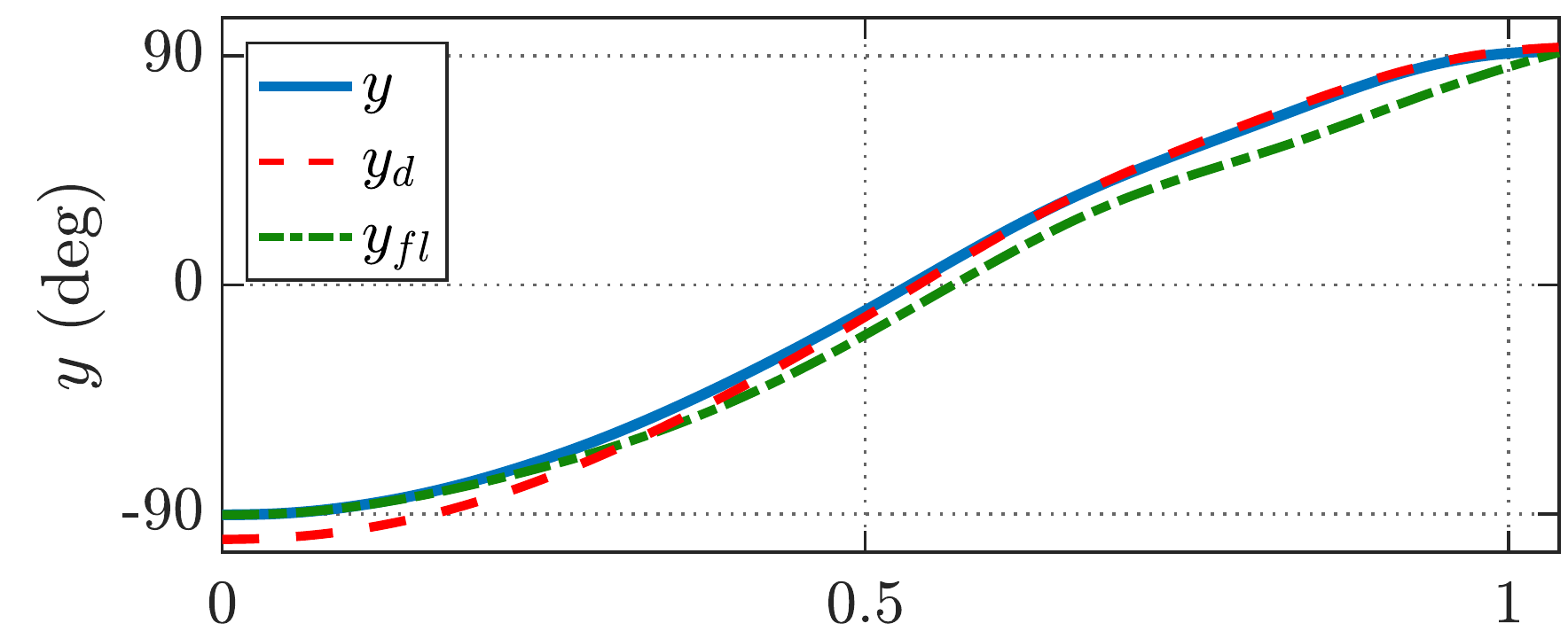}} & 
{\includegraphics[trim={1bp 105bp 50bp 110bp},clip,width=0.5\columnwidth]{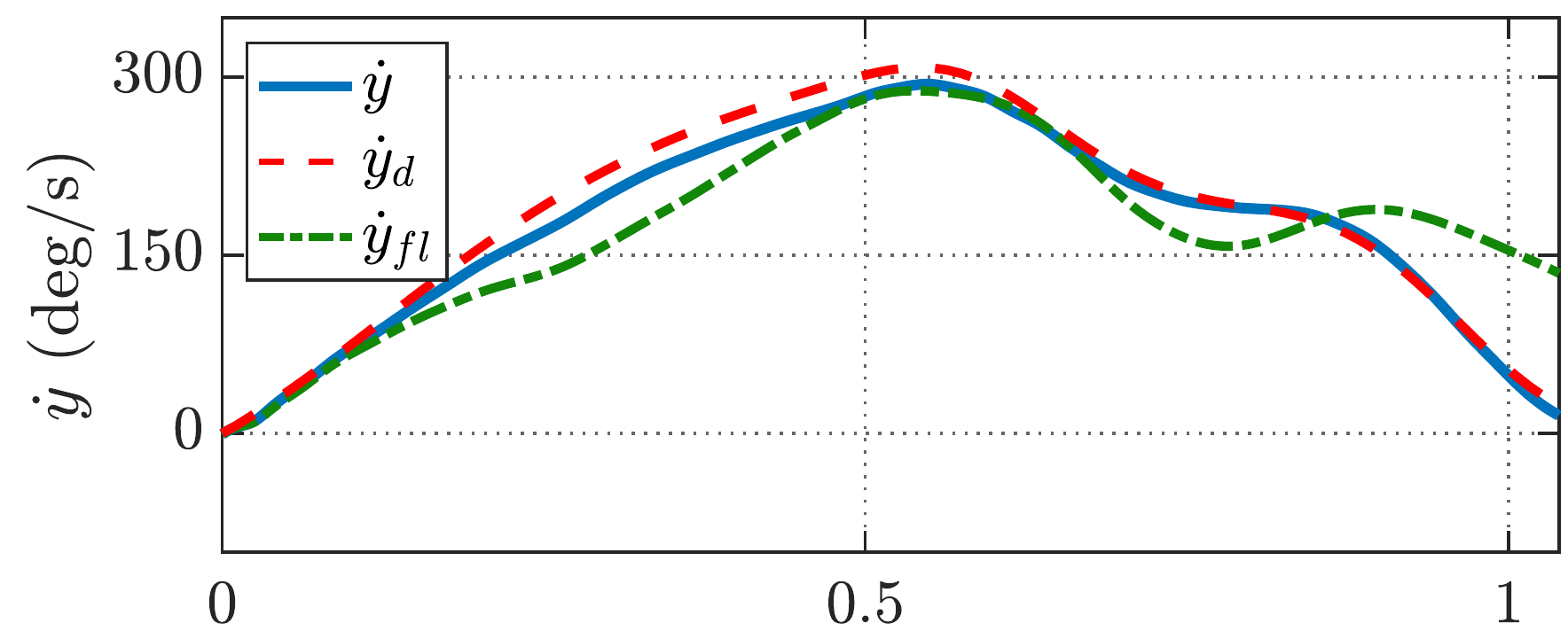}}
\tabularnewline
{\scriptsize{}(c)} & {\scriptsize{}(d)}
\tabularnewline
{\includegraphics[trim={5bp 7bp 50bp 18bp},clip,width=0.5\columnwidth]{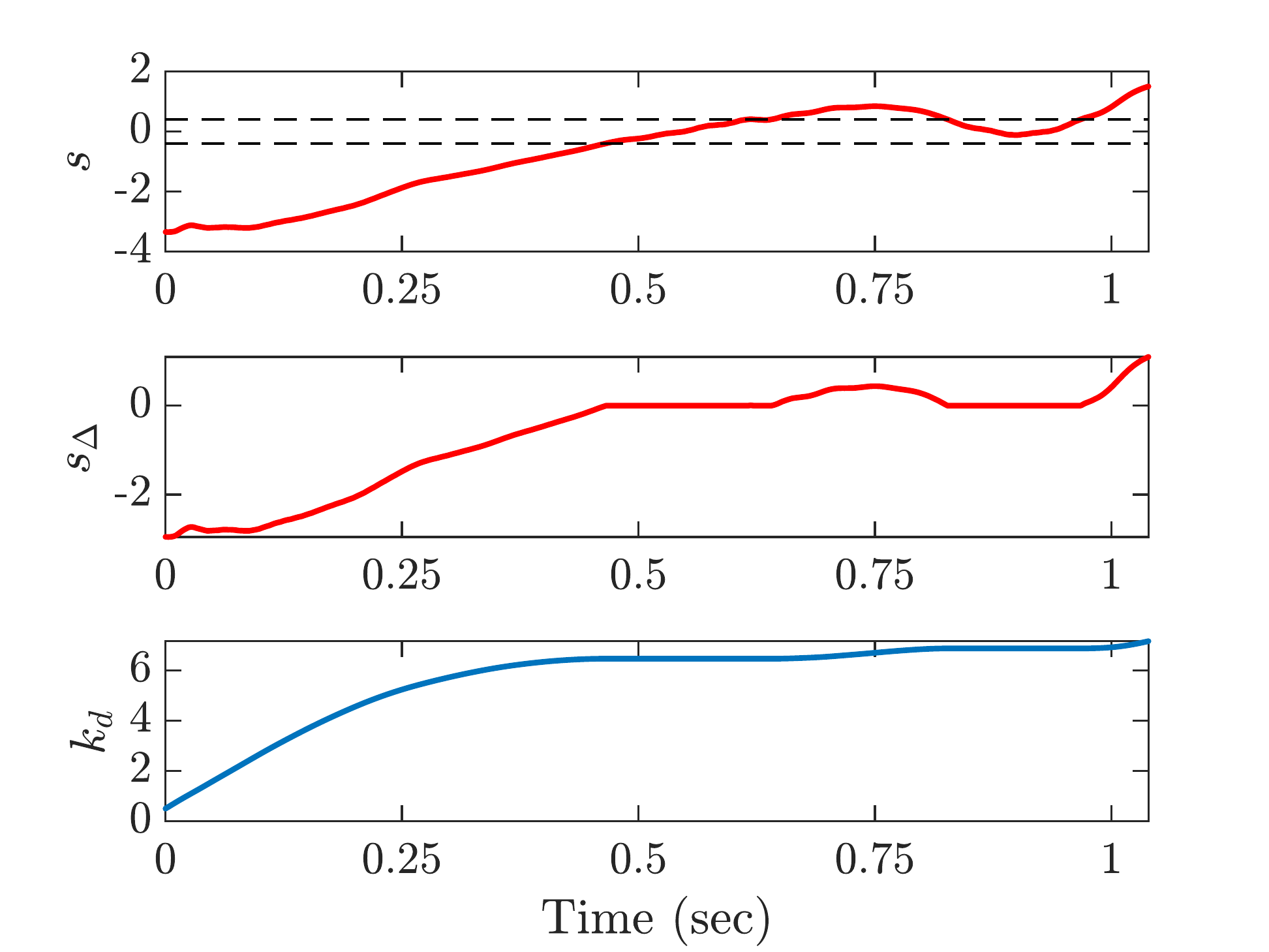}} & 
{\includegraphics[trim={5bp 7bp 50bp 18bp},clip,width=0.5\columnwidth]{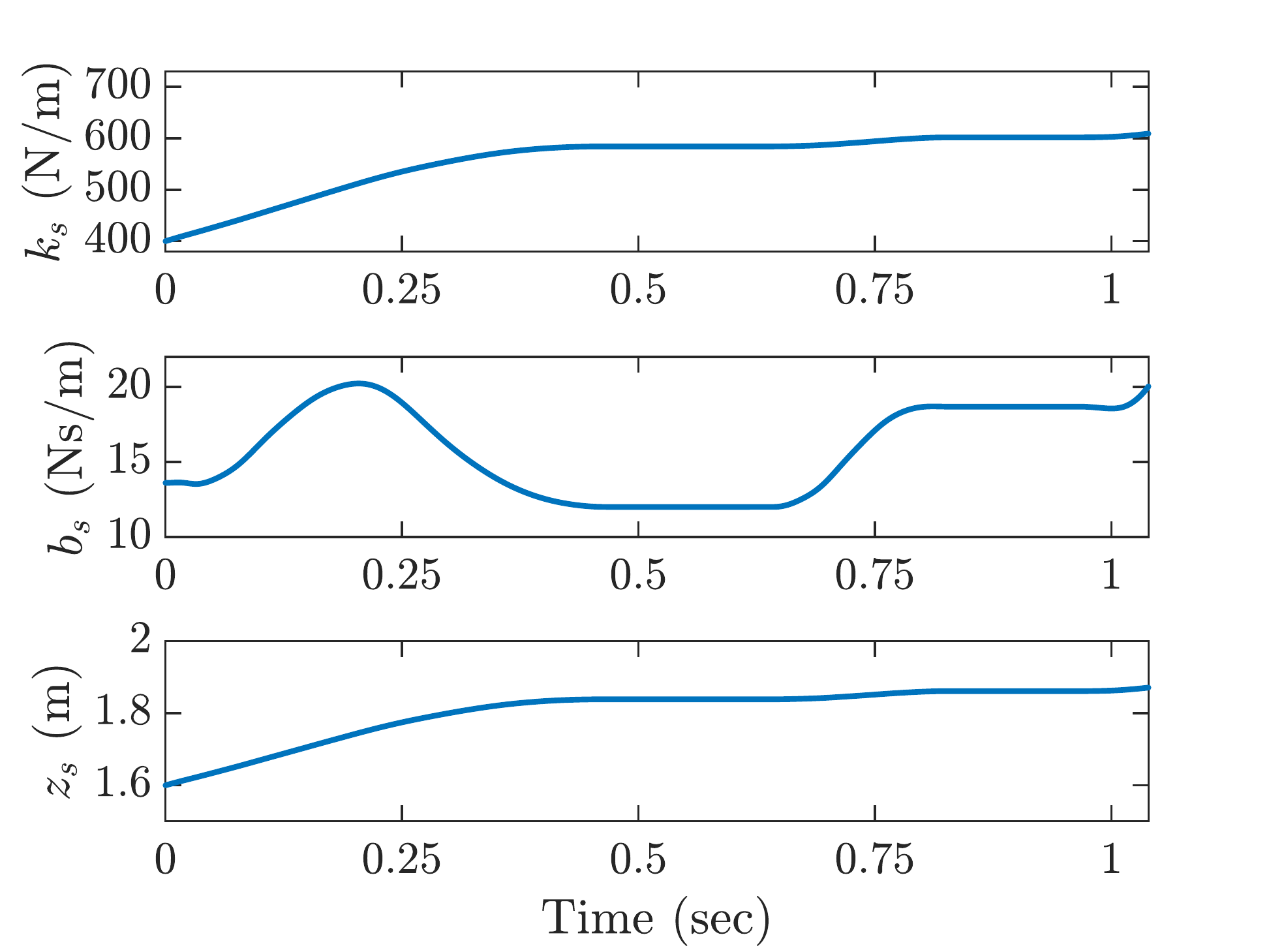}}
\tabularnewline
{\scriptsize{}(e)} & {\scriptsize{}(f)}
\tabularnewline
\multicolumn{2}{c}{\includegraphics[trim={5bp 80bp 50bp 110bp},clip,width=0.53\columnwidth]{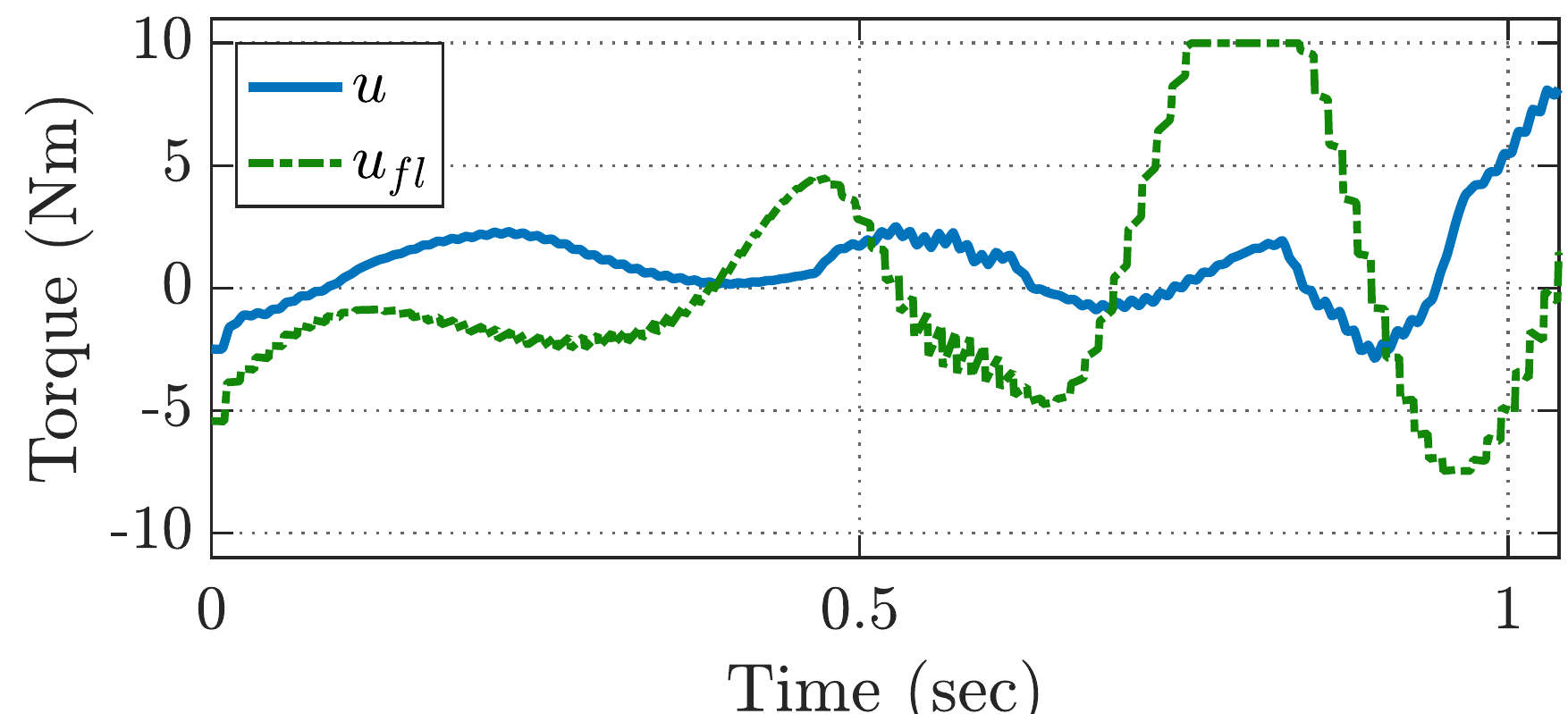}}
\tabularnewline
\multicolumn{2}{c}{\scriptsize{}(g)}
\end{tabular}
\caption{\label{fig:simulation-cable} Brachiation on an unknown flexible cable, starting from off-nominal initial configuration: (a) motion trajectory under the proposed controller, (b) motion trajectory under the baseline controller, (c, d) output trajectories for the proposed and baseline, (e) boundary layer trajectories and the robust gain $k_d$ (the black dashed lines represent the boundary layer between $\pm\phi$), (f) estimated parameters $\hat{p}$, (g) torque profiles for the proposed and baseline.}
\end{figure}

\vspace{-2pt}
We first validate the performance of the proposed controller for the scenario in which the robot starts on the cable with unknown dynamics from the off-nominal initial condition of $[-35^{\circ},-110^{\circ},1.84\,\textrm{m},0,0,0]$. The results of the proposed and the baseline controllers are compared in Figs. \ref{fig:simulation-cable}(a)-(d).\looseness=-1

As illustrated in Fig. \ref{fig:simulation-cable}(a),
the proposed adaptive robust controller successfully tracks the desired trajectory and drives the robot to the desired configuration, exhibiting robustness to the unmodeled dynamics and disturbances caused by the flexible cable dynamics. Although the baseline controller also eventually gets the robot to reach the cable (Fig. \ref{fig:simulation-cable}(b)), the final configuration of the robot is different from the desired configuration, and the final joint velocities are significantly high, resulting in the robot crashing into the cable. These issues are caused by the fact that the baseline controller cannot accurately track the desired output trajectories during the swing motion due to the uncertainties and disturbances present in the system, as will be discussed below. Note that for a successful brachiation motion, the robot is required to traverse a minimum desired distance along the cable, hence the importance of the desired final configuration.
Figs. \ref{fig:simulation-cable}(c)-(d) show the output tracking performance of the proposed controller compared to the baseline. While the output of the system under the proposed controller converges to and tracks the desired trajectory (with $\textrm{RMSE}_y{=}3.2^{\circ}$ and $\textrm{RMSE}_{\dot{y}}{=}14.7$ deg/s), the output trajectories under the baseline controller violate the desired trajectories (with $\textrm{RMSE}_y{=}8.9^{\circ}$ and $\textrm{RMSE}_{\dot{y}}{=}43.4$ deg/s) due to the uncertainties and disturbances caused by the flexible cable.\looseness=-1

Figs. \ref{fig:simulation-cable}(e)-(f) show the estimated parameter vector $\hat{p}$ as well as the evolution of the robust gain $k_d$.
The trajectories $s$ and $s_{\Delta}$ are also plotted in Fig. \ref{fig:simulation-cable}(e). Based on the value of $\phi$ and the definition of $s_{\Delta}$, it is seen that the direct and indirect adaptation mechanisms are active only when $s$ is outside the boundary layer (i.e., $s_{\Delta}$ is nonzero). When $s_{\Delta}$ is zero, the parameter/gain adaptations stop and the estimated parameters $\hat{p}$ and the robust gain $k_d$ remain constant.
As expected, for the first part of the swing, when the estimated parameter errors are large, the $s$ trajectory exceeds the boundary later (the area between $\phi=-0.4$ and $\phi=0.4$) and the $s_{\Delta}$ trajectory is non-zero. Hence, the robust controller gain $k_d$ updates to compensate for the disturbance (Fig. \ref{fig:simulation-cable}(e)). Similarly, the estimated parameters evolve over time and converge to their final values (Fig. \ref{fig:simulation-cable}(f)). Better parameter estimation by the proposed controller results in better system modeling so that the controller achieves a better tracking performance.
Although we cannot verify the accuracy of the estimated parameters for this experiment (since the true physical parameters of the spring-damper model equivalent to the flexible cable dynamics are unknown),
any discrepancy between the estimated parameters and the true equivalent parameters is considered as a disturbance, compensated by the robust control term.
We validate the accuracy of the parameter estimation in the next section by performing brachiation on a spring-damper model.\looseness=-1

Fig. \ref{fig:simulation-cable}(g) compares the control torque inputs for the proposed and the baseline controllers. The torque applied by the proposed controller is in the range of $[-3,\,8]$ N\,m, which is well within the saturation limits of $\pm10$ N\,m and is relatively optimal compared to the torque applied by the baseline controller ($\textrm{RMS}_{u}=2.0$ vs. $\textrm{RMS}_{u_{fl}}=4.5$ N\,m).

\subsection{Validation of the Adaptive Cable Force Estimation and Robustness to the Residual Force and Disturbances}
\begin{figure}[t]
\renewcommand{\arraystretch}{0.2}
\setlength\tabcolsep{0.5pt}
\begin{tabular}{cc}
{\begin{tikzpicture}\node[inner sep=0pt] at (0,0)
{\includegraphics[trim={5bp 1bp 50bp 20bp}, clip, width=0.48\columnwidth] {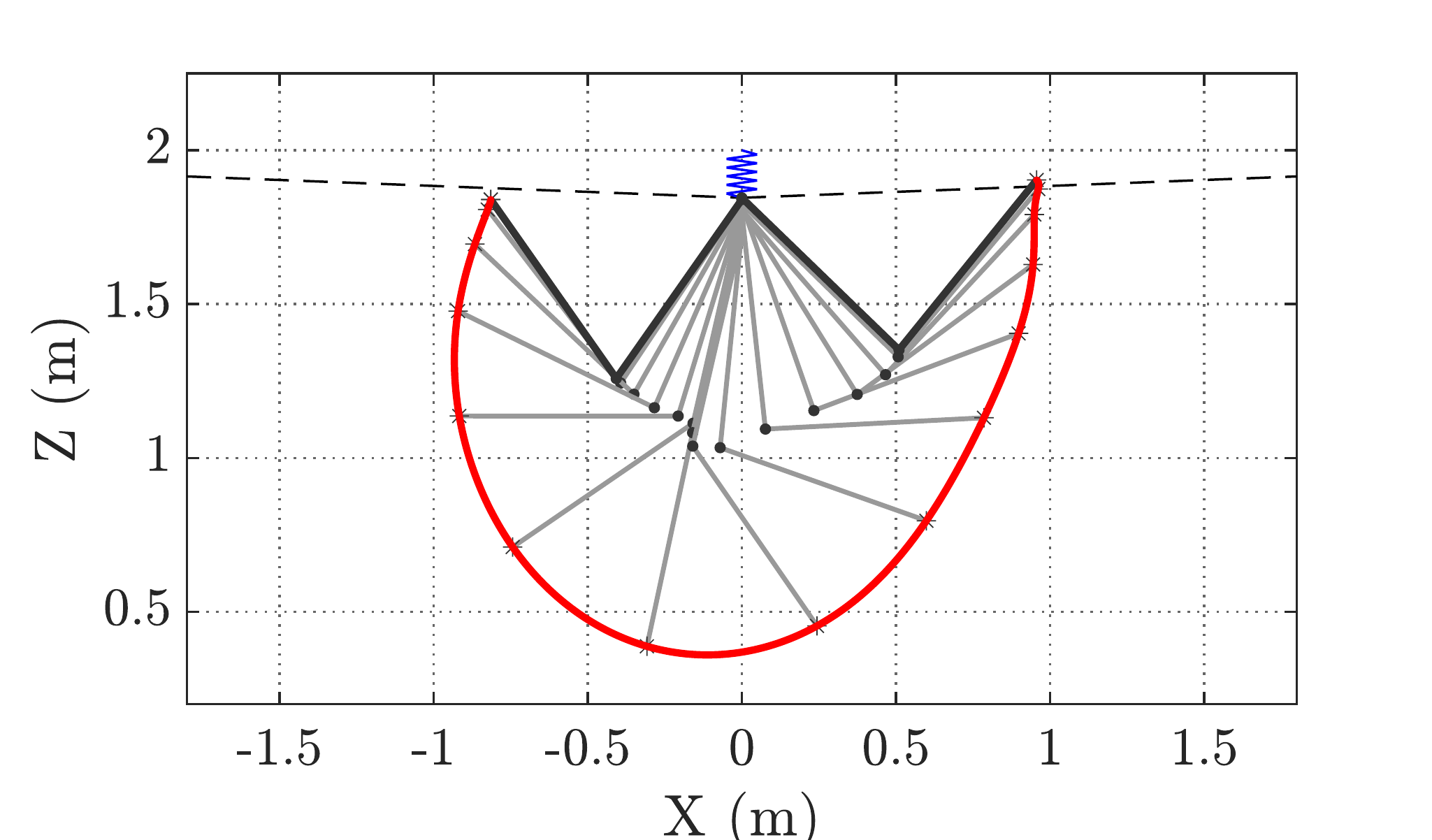}};
\draw[<-,>=stealth',semithick,red,dashed] (-0.75,-0.1) to [out=130,in=230] (-0.75,0.75);
\end{tikzpicture}} & 
{\begin{tikzpicture}\node[inner sep=0pt] at (0,0)
{\includegraphics[trim={5bp 1bp 50bp 20bp}, clip, width=0.48\columnwidth] {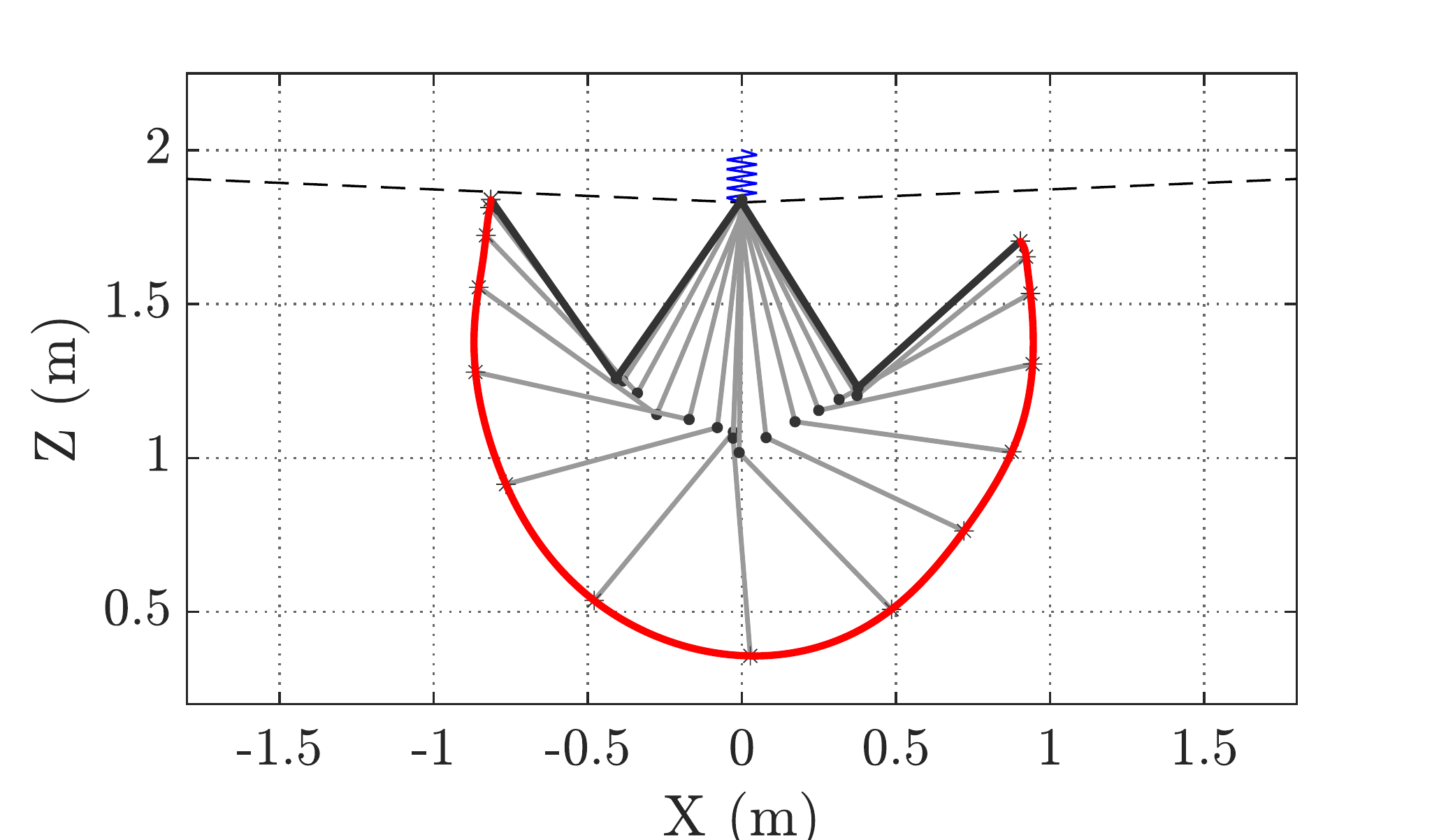}};
\draw[<-,>=stealth',semithick,red,dashed] (-0.75,-0.1) to [out=130,in=230] (-0.75,0.75);
\end{tikzpicture}}
\tabularnewline
{\scriptsize{}(a)} & {\scriptsize{}(b)}
\tabularnewline
{\includegraphics[trim={1bp 105bp 50bp 110bp},clip,width=0.5\columnwidth]{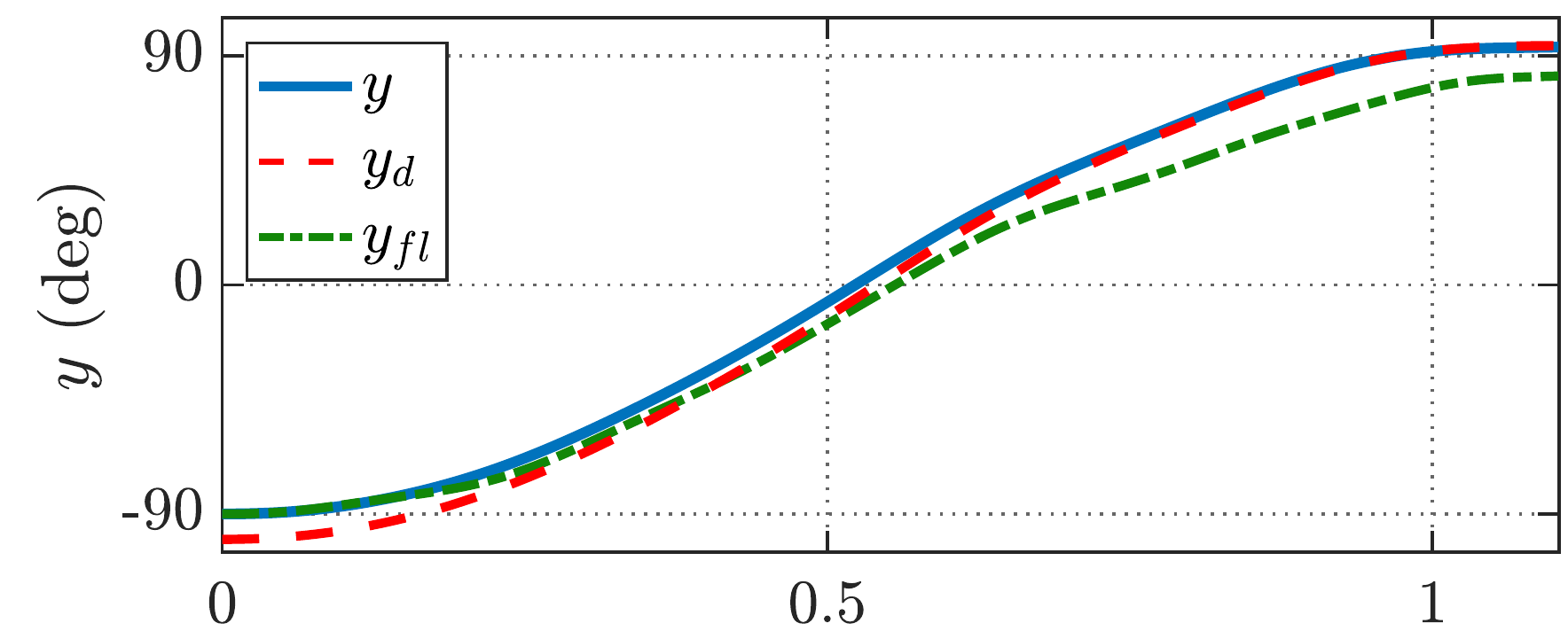}} & 
{\includegraphics[trim={1bp 105bp 50bp 110bp},clip,width=0.5\columnwidth]{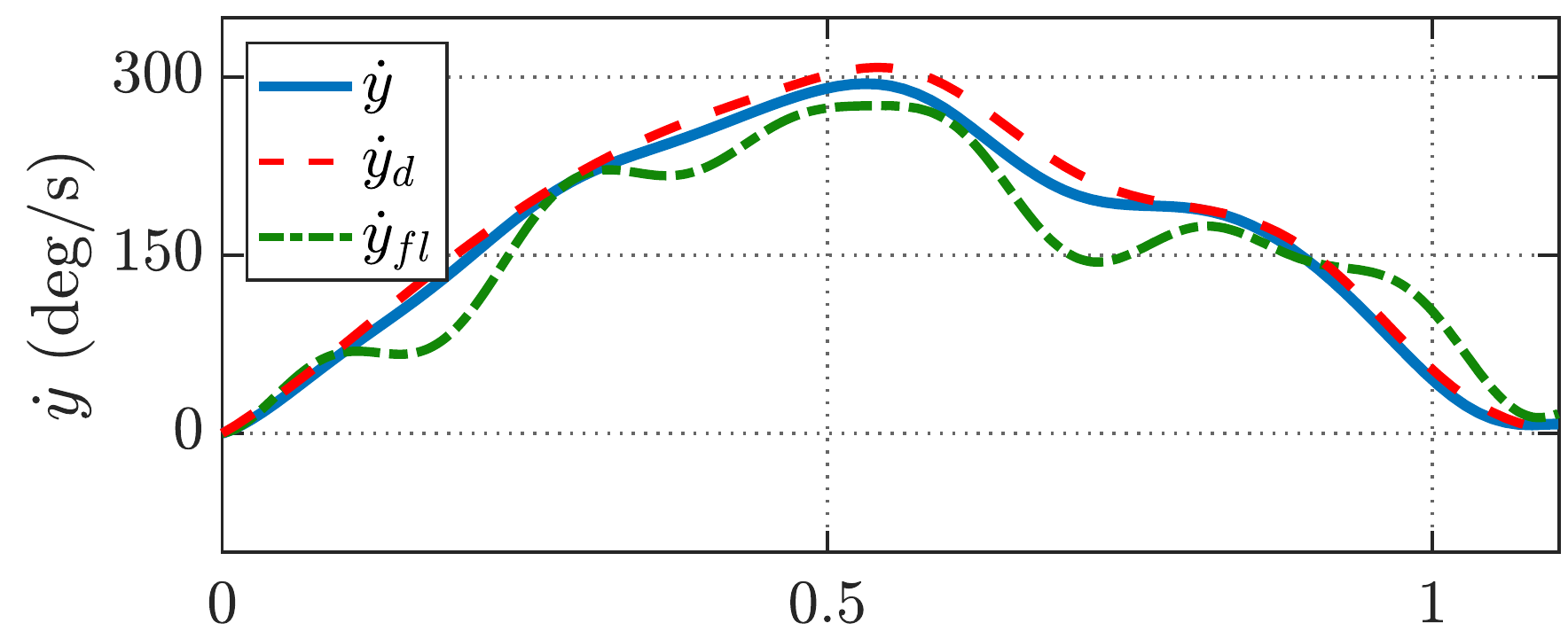}}
\tabularnewline
{\scriptsize{}(c)} & {\scriptsize{}(d)}
\tabularnewline
{\includegraphics[trim={5bp 28bp 50bp 18bp},clip,width=0.5\columnwidth]{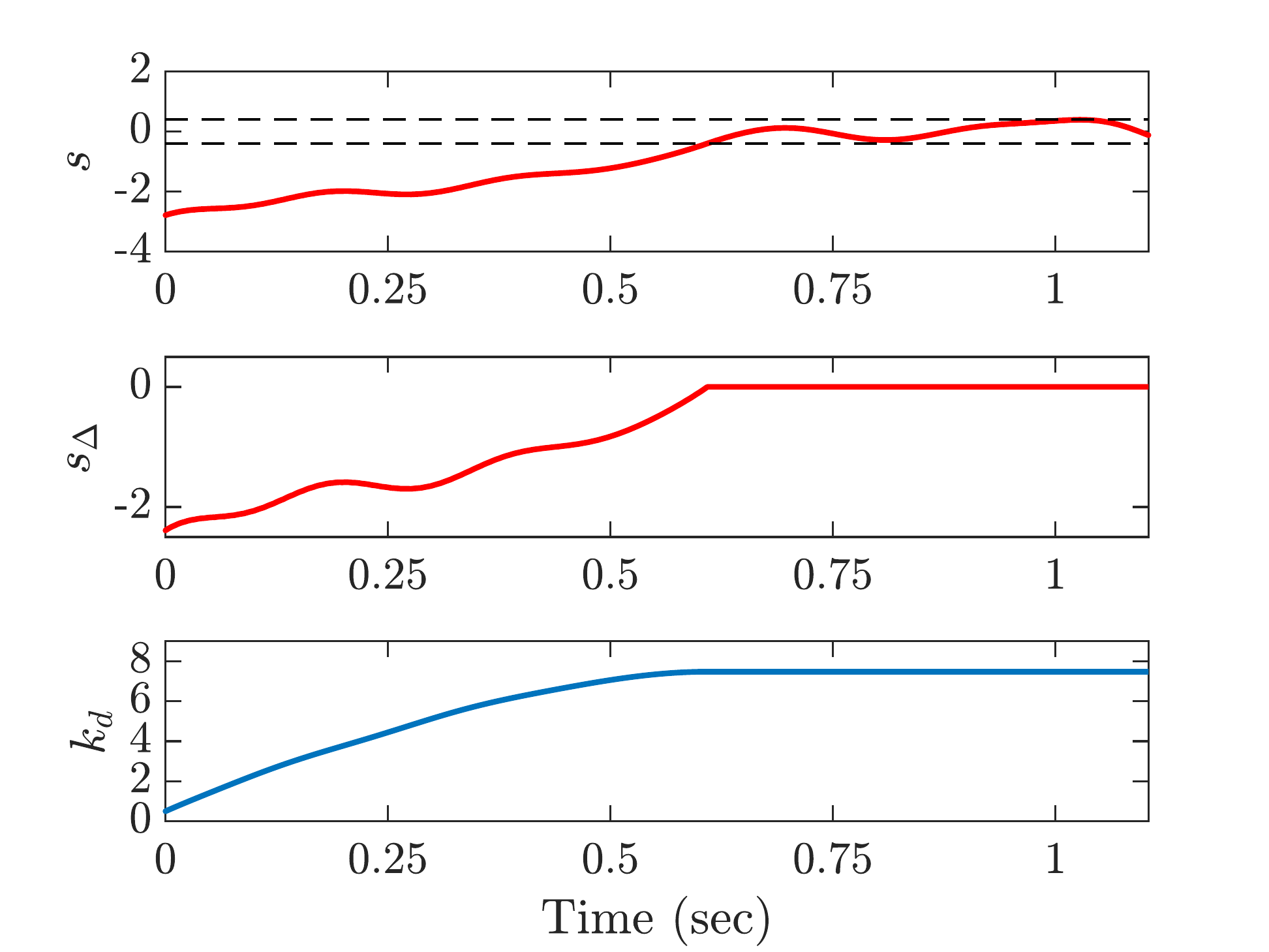}} & 
{\includegraphics[trim={5bp 28bp 50bp 18bp},clip,width=0.5\columnwidth]{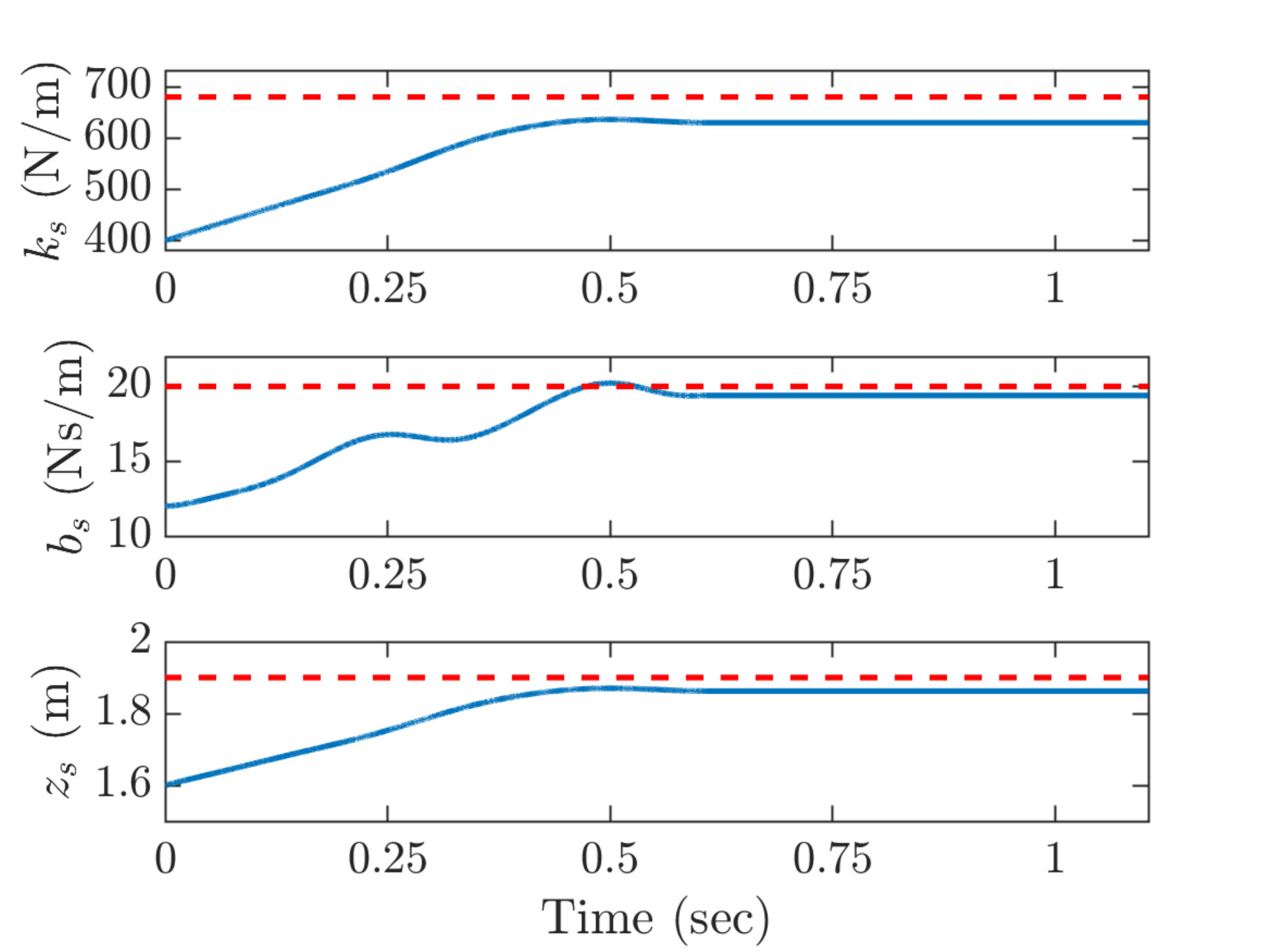}}
\tabularnewline
{\scriptsize{}(e)} & {\scriptsize{}(f)}
\tabularnewline
{\includegraphics[trim={5bp 80bp 50bp 110bp},clip,width=0.5\columnwidth]{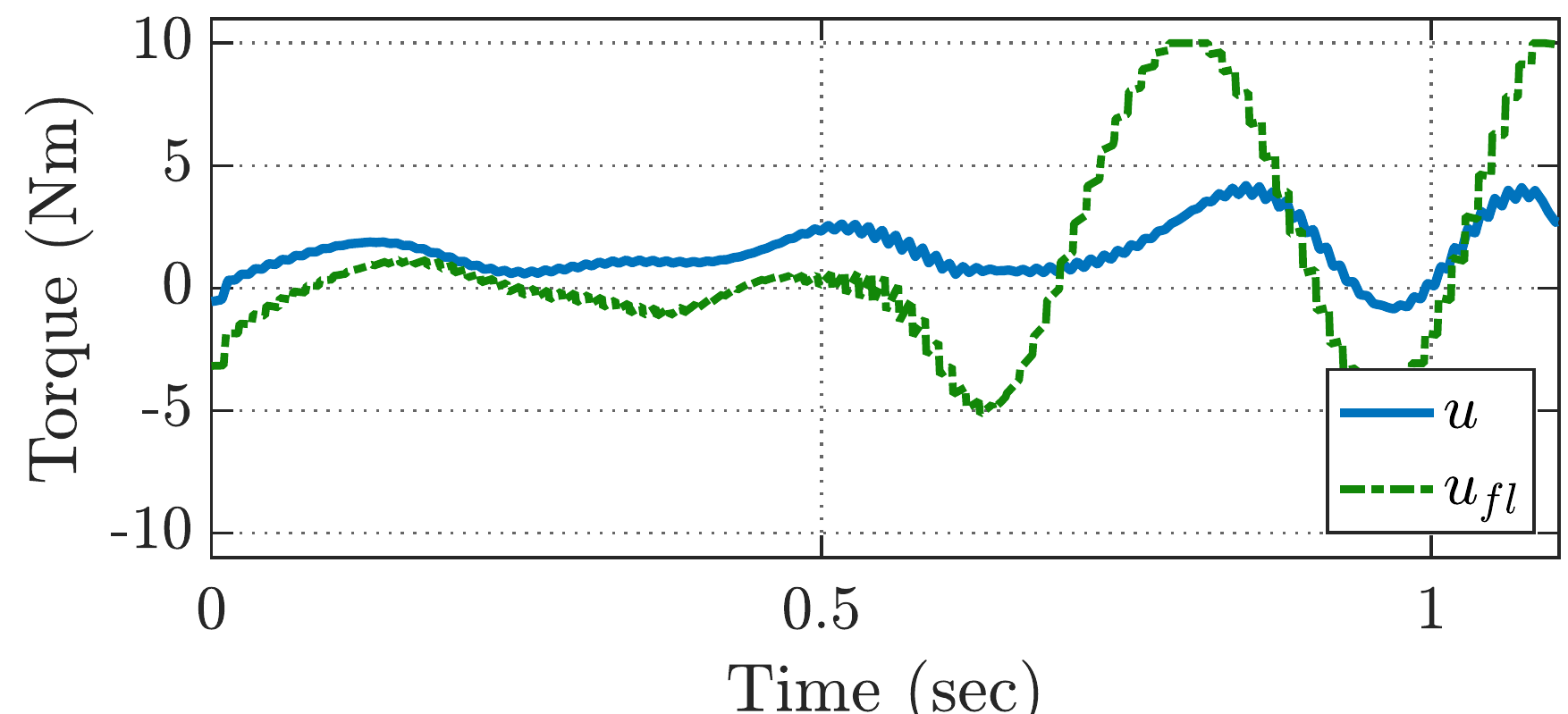}} &
{\includegraphics[trim={5bp 80bp 50bp 105bp},clip,width=0.5\columnwidth]{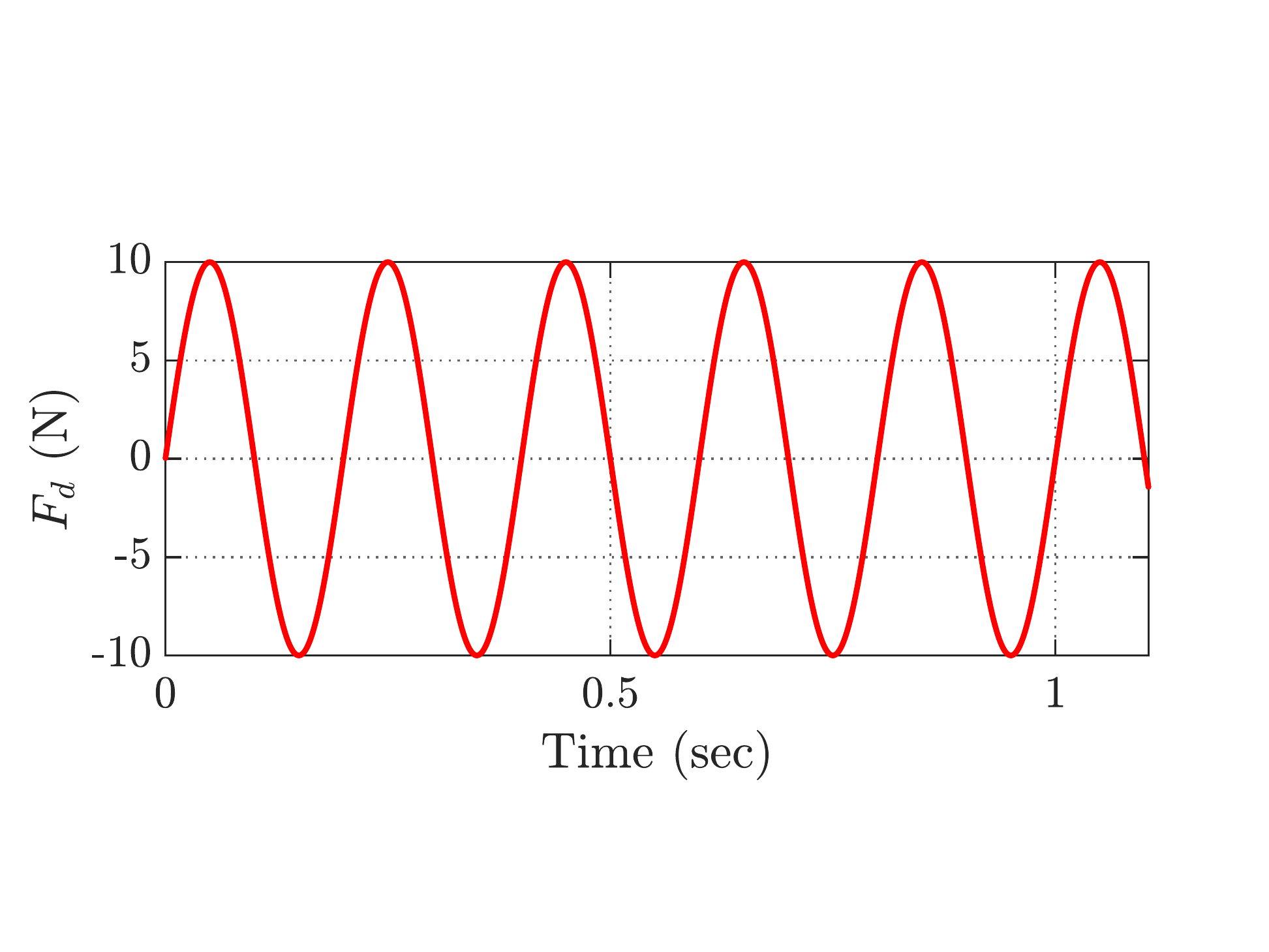}}
\tabularnewline
{\scriptsize{}(g)} & {\scriptsize{}(h)}
\end{tabular}
\caption{\label{fig:simulation-spring} Brachiation on an unknown spring-damper with a time-varying external disturbance, starting from off-nominal initial configurations: (a) motion trajectory under the proposed controller, (b) motion trajectory under the baseline controller, (c, d) output trajectories for the proposed and baseline controllers, (e) boundary layer trajectories and the robust gain $k_d$ (the black dashed lines represent the boundary layer between $\pm\phi$), (f) estimated parameters $\hat{p}$ (dashed red lines represent the true parameters), (g) torque profiles for the proposed and baseline controllers, (h) external disturbance $F_d$ applied to the pivot gripper.}
\end{figure}

\vspace{-2pt}
The experiment for brachiating on the full-cable model
presented in the previous section resembles an actual scenario,
where the cable's dynamic model and its force applied to the
robot are unknown. In order to evaluate the performance of
the indirect adaptation mechanism to estimate the equivalent spring-damper parameters, we conduct a simulation on the low-fidelity model, where the pivot gripper is attached to a spring-damper instead of the full flexible cable.
The spring-damper dynamic model and
its parameters are again unknown to the controller. Moreover, to
substitute
for the residual external forces applied by an actual cable, we apply a time-varying external disturbance to the pivot gripper, described by $F_d{=}10 \sin{(10\pi t)}$, as plotted in Fig. \ref{fig:simulation-spring}(h). A sinusoidal disturbance is chosen to replicate other harmonics generated by the cable, not captured by the single spring-damper.
For this simulation, the actual spring-damper parameters are chosen as: $k_s=680$ N/m, $b_s=20$ N\,s/m and $z_s=1.9$ m.\looseness=-1

The results for brachiation of the robot attached to the spring-damper model are presented in Fig. \ref{fig:simulation-spring}. Under the proposed controller, both angle and velocity of the output $y$ accurately converge to and track the desired trajectories and the robot reaches a ``virtual'' cable. However, the robot performance under the baseline controller is unsuccessful in tracking the desired trajectory and reaching the cable, due to the uncertainty and disturbances present in the system.

As shown in Fig. \ref{fig:simulation-spring}(f), the adaptive  parameters estimated by the indirect adaptation scheme approximately converge to the actual spring-damper parameters. However, as expected, the estimation error is not zero due to the short time horizon of the swing motion, nevertheless the robust term enables the controller to maintain the tracking and stability in the presence of such bounded uncertainty. 
Moreover, the successful performance of the robot and its invariance to the applied external disturbance demonstrate the effectiveness of the proposed robust control design included in the control law.\looseness=-1

\subsection{Monte Carlo Simulation for a Range of Initial Conditions} \label{subsec:monte-carlo}
\begin{figure}[t]
\renewcommand{\arraystretch}{0.4}
\setlength\tabcolsep{0.5pt}
\includegraphics[trim={90bp 0bp 110bp 20bp}, clip,width=\columnwidth]{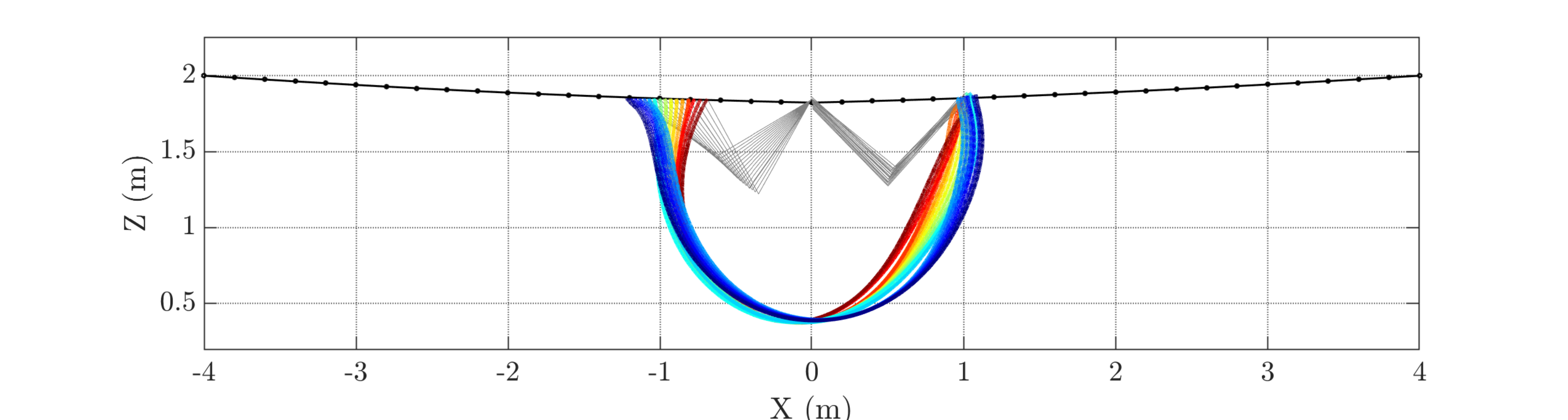}
\vspace{-18pt}
\caption{\label{fig:monte-carlo} Monte Carlo simulation consisting of 20 runs of the proposed adaptive robust controller for the robot starting from different initial configurations on an uknown flexible cable.}
\end{figure}

\vspace{-2pt}
A Monte Carlo simulation is carried out to further elaborate on the reliability of the proposed adaptive robust controller for the robot starting from different initial conditions on an unknown cable.
Using a single desired output trajectory, we performed 20 simulations, each starting from a different initial condition on the cable in the range of ${-60^{\circ}}{<}\theta_1{<}{-30^{\circ}}$ and ${-120^{\circ}}{<}\theta_2{<}{-60^{\circ}}$.
Fig. \ref{fig:monte-carlo} shows the resultant motion trajectories. As can be seen on the plots, for all the cases, the proposed adaptive robust controller successfully drives the robot to the desired configuration and reach the cable.

\begin{figure}
\renewcommand{\arraystretch}{0.4}
\setlength\tabcolsep{1pt}
\centering
\vspace{-10pt}
\begin{tabular}{cc}
{\includegraphics[trim={6bp 45bp 50bp 85bp},clip,width=0.45\columnwidth]{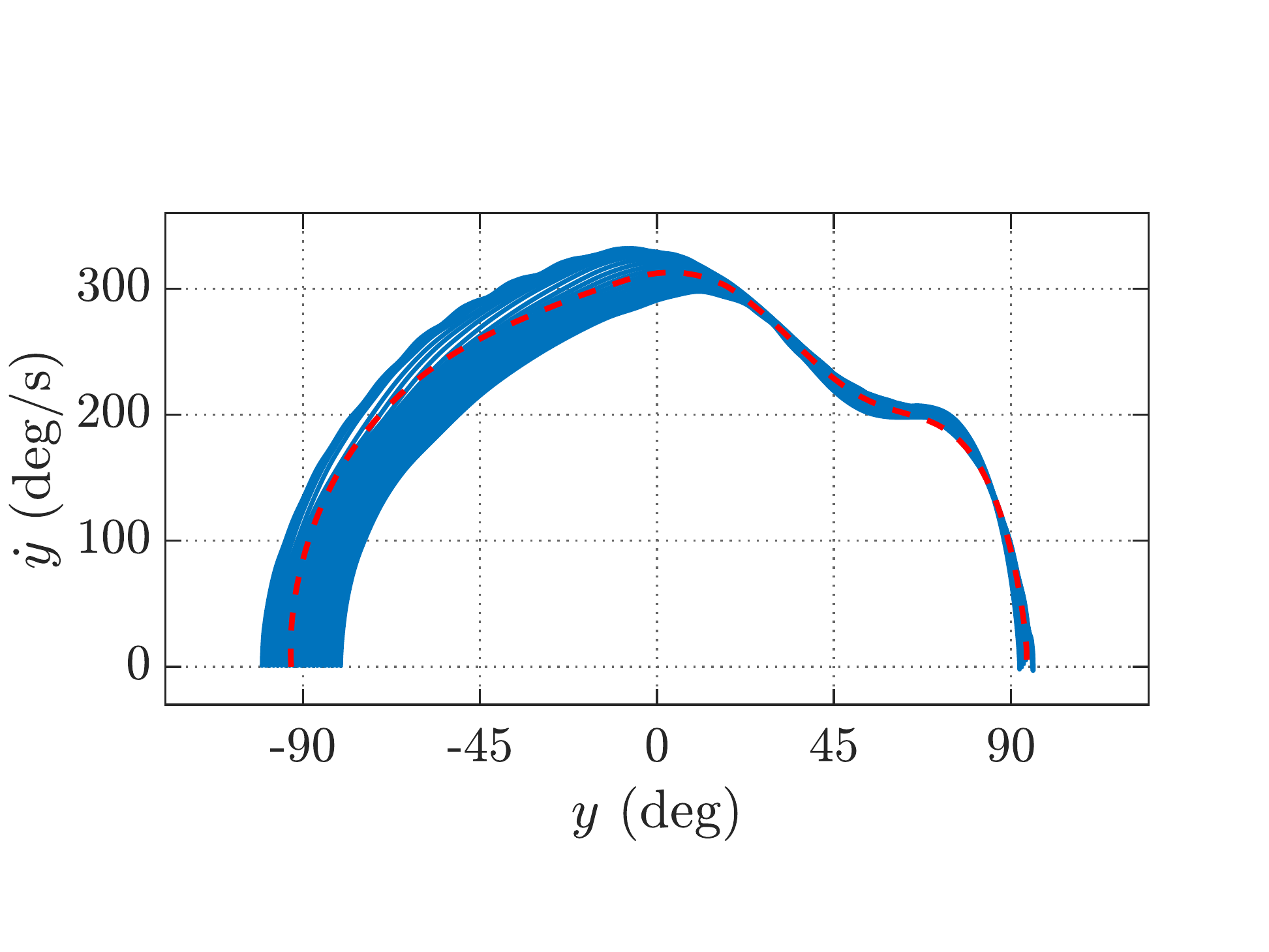}} & 
{\includegraphics[trim={6bp 45bp 50bp 85bp},clip,width=0.45\columnwidth]{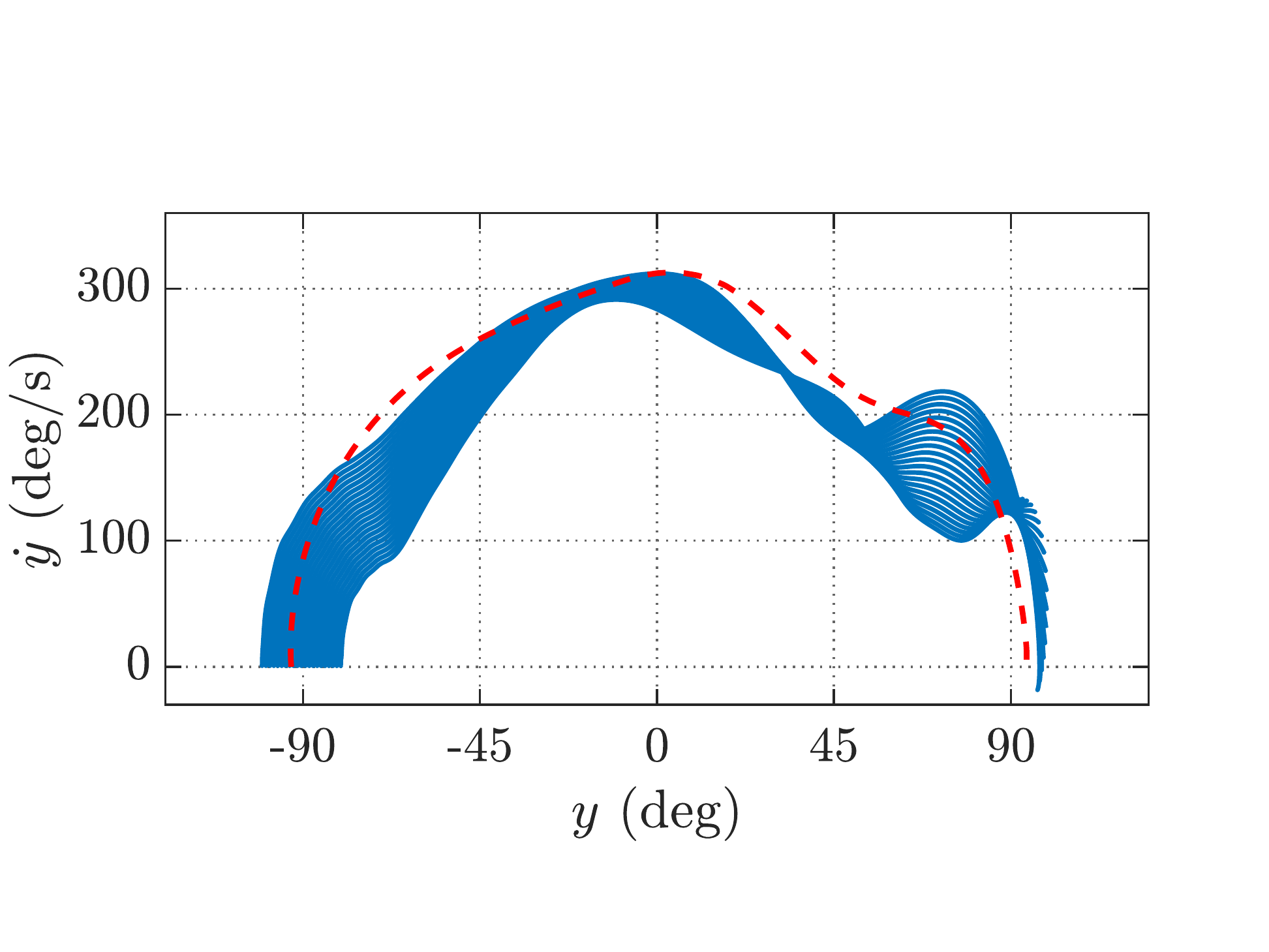}} \tabularnewline
{\scriptsize{}(a)} & {\scriptsize{}(b)} \tabularnewline
\end{tabular}\caption{\label{fig:phase-port}Phase portrait of the output angle and velocity for brachiation starting from different initial conditions using (a) the proposed adaptive robust controller, and (b) the baseline input-output feedback linearization method. The red dashed line indicates the desired output trajectory.}
\end{figure}

Fig. \ref{fig:phase-port} illustrates the phase portraits for 20 swings performed starting from different initial conditions on an unknown cable, using both the baseline and the proposed controllers. As is evident on the plots, the proposed control design maintains a consistent position-velocity tracking performance compared to the baseline method, successfully achieving the final desired output angle and velocity. The numerical RMS results for the two controllers are listed in Table II. The performance of the proposed adaptive robust controller dominates that of the feedback linearization, as it improves the control optimality ($\textrm{RMS}_u$) by $23\%$, and the tracking errors ($\textrm{RMSE}_y$ and $\textrm{RMSE}_{\dot{y}}$) by $45\%$ and $75\%$ respectively. The results indicate the reliability and optimality of the proposed control design and emphasize the importance of designing a robust and adaptive control scheme for such applications.\looseness=-1

\begin{table}[h!]
    \vspace{-5pt}
    \caption{Numerical results compared to the baseline controller}
    \label{tab:result-compare}
    \centering
    \begin{tabular}{cccc}
    \toprule
     Method & $\textrm{RMS}_u$ (Nm) & $\textrm{RMSE}_y$ (deg) & $\textrm{RMSE}_{\dot{y}}$ (deg/s) \\
    \midrule
    Feedback Lin. & 5.08 & 7.83 & 42.84 \\
    Adaptive Robust & 3.92 & 4.29 & 10.55 \\
    \bottomrule
    \end{tabular}
    \vspace{-10pt}
\end{table}

\subsection{Continuous Brachiation over a Full-Cable Model} \label{subsec:continuous-sim}
\vspace{-2pt}
In order for the robot to traverse the entire length of the cable, it is required to perform ``continuous'' sequential swings. For this scenario, the cable continues to vibrate significantly throughout the robot motion and the robot starts from non-zero dynamic states for all except the first swing, which make achieving such
brachiating locomotion a challenge.

\begin{figure}
\renewcommand{\arraystretch}{0.25}
\centering
\begin{tabular}{c}
\includegraphics[trim={125bp 30bp 130bp 70bp}, clip,width=0.9\columnwidth]{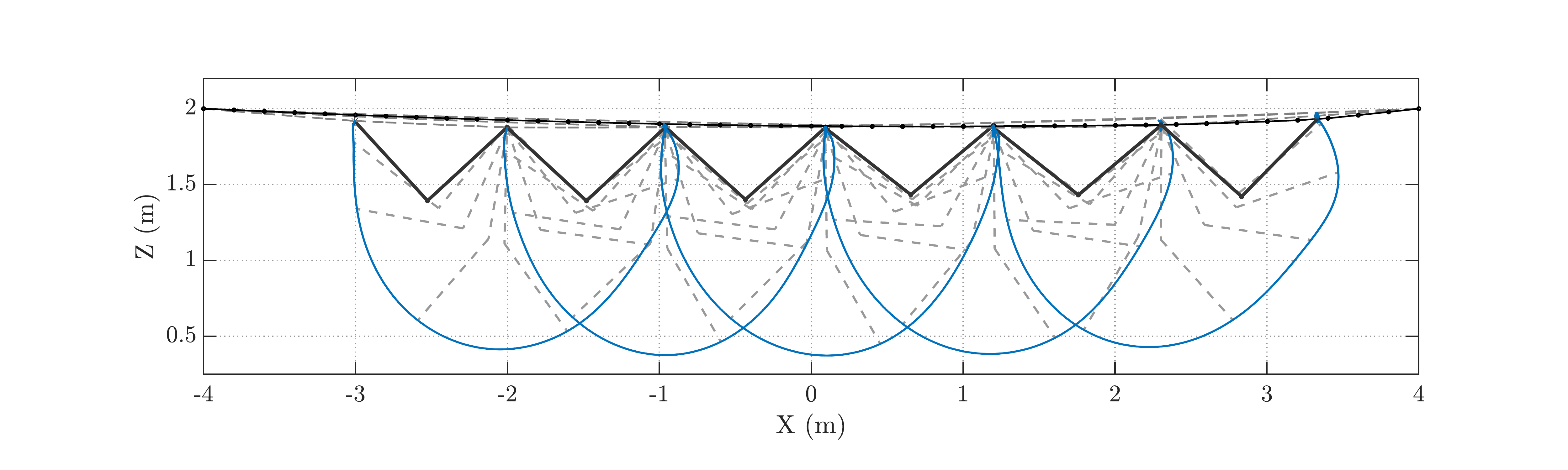}
\tabularnewline
{\scriptsize{}(a)}
\tabularnewline
\includegraphics[trim={90bp 41bp 110bp 18bp}, clip,width=0.89\columnwidth]{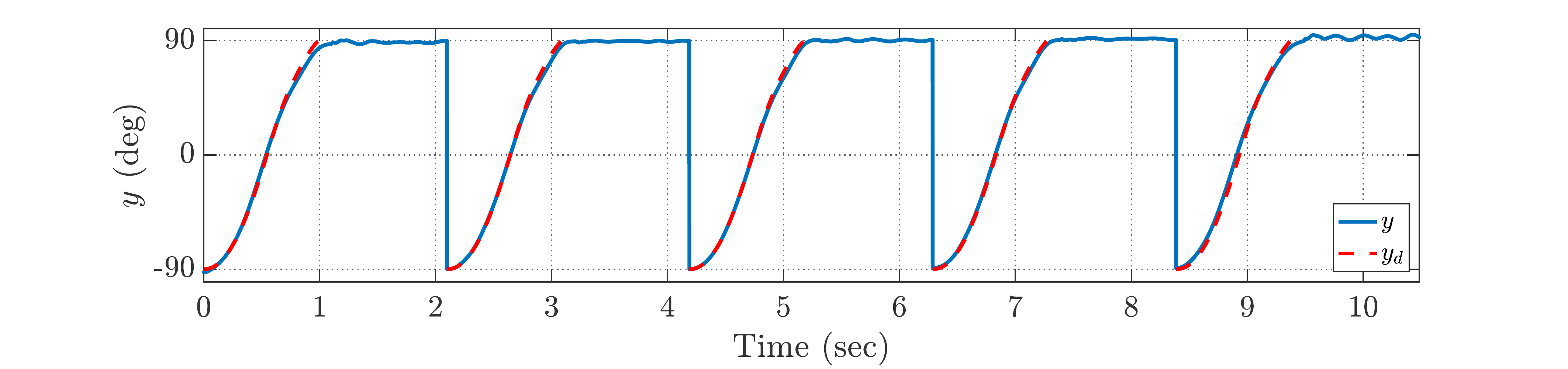}
\tabularnewline
{\scriptsize{}(b)}
\tabularnewline
\includegraphics[trim={86bp 41bp 110bp 18bp}, clip,width=0.89\columnwidth]{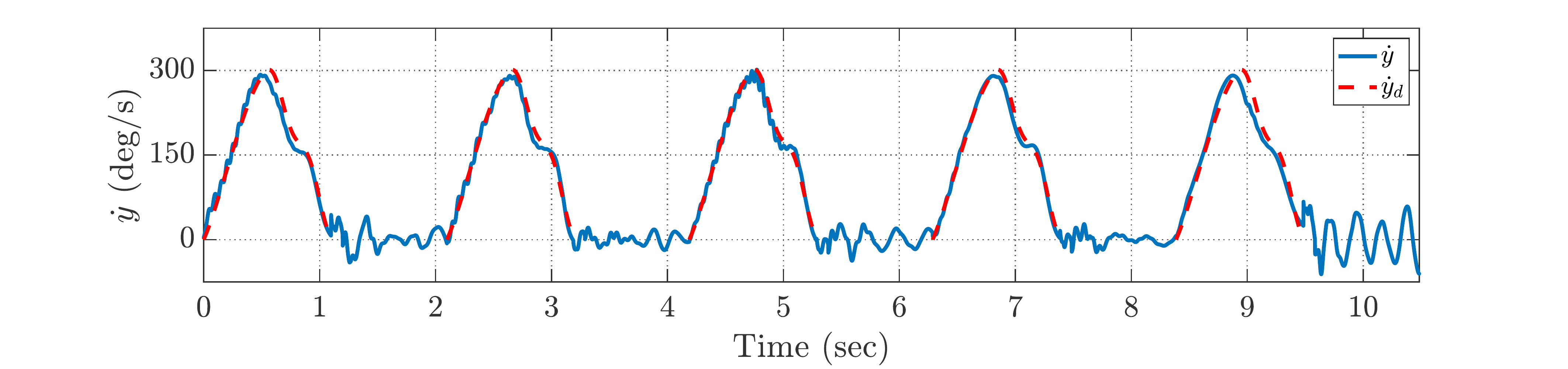}
\tabularnewline
{\scriptsize{}(c)}
\tabularnewline
\includegraphics[trim={90bp 5bp 110bp 12bp}, clip,width=0.89\columnwidth]{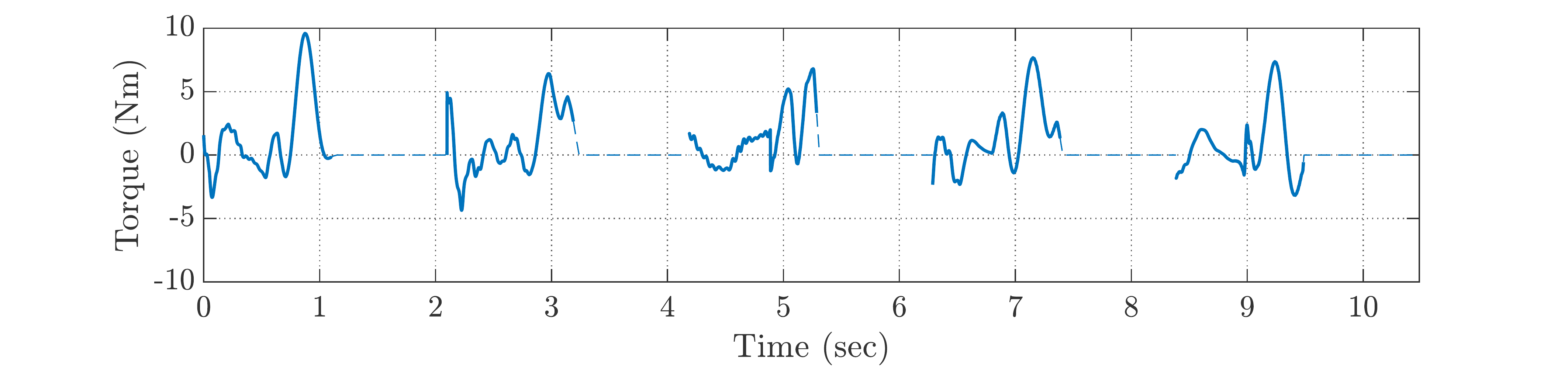}
\tabularnewline
{\scriptsize{}(d)}
\tabularnewline
\end{tabular}
\caption{\label{fig:continuous-sim} Continuous brachiation on an unknown flexible cable using a single desired trajectory under the proposed controller: (a) motion trajectories, (b) output angle, (c) output velocity, (d) torque profile.}
\vspace{-4pt}
\end{figure}

Fig. \ref{fig:continuous-sim} evaluates the performance of the proposed controller for the continuous brachiation scenario. 
The robot starts on an unknown flexible cable with unmodeled dynamics from the initial conditions of $[-46.2^{\circ},-89.1^{\circ},1.88\,\textrm{m},0,0,0]$. The control input is constrained by the torque limits of $\pm 10$ N\,m, and the spring-damper parameters are again initialized to $k_s=400$ N/m, $b_s=12$ N\,s/m, and $z_s=1.6$ m.

While the reference trajectory is not designed for the exact
configuration of each swing, the proposed adaptive-robust control framework enables the robot to successfully traverse the entire length of the cable in 5 swings, all in the presence of unmodeled dynamics, estimation uncertainties, cable disturbances and control saturations. Note that a pause of only $1$ second is enforced between sequential swings, so that the cable is heavily oscillating when starting each swing. 

As shown in Figs. \ref{fig:continuous-sim}(a)-(c), for all the swings, the output of the system under the proposed control accurately tracks the desired trajectory and the swing gripper reaches the cable with the desired configuration. The minor error in the velocity ($\dot{y}$) tracking performance is due to the large scaling factor $\lambda$ chosen for the controller, as tracking the output angle has more importance than the velocity for this application.

The torque input generated by the controller is maintained within the torque limits of $\pm10$ N\,m (Fig. \ref{fig:continuous-sim}(d)). The adaptive estimated spring-damper parameters follow the same trend as the results presented in Fig. \ref{fig:simulation-cable}, and are not plotted again due to space limitations. The superior performance of the proposed controller also demonstrate that the bounded estimation errors and any residual forces (i.e. disturbances) applied by the cable are appropriately handled by the robust control design.\looseness=-1

\section{CONCLUSIONS}
\vspace{-2pt}
We synthesized a combined direct-indirect adaptive robust control framework for underactuated two-link brachiating robots traversing flexible cables. A novel low-fidelity dynamic model for the robot-cable system was proposed, in which the dynamic effects of the cable on the robot are modeled as a spring-damper force plus a residual external force applied to the pivot gripper. The proposed model provided the ability to include the cable as an unmodeled dynamics with parametric uncertainties and non-parametric disturbances in the system. An indirect adaptive scheme was developed to estimate the physical parameters of the cable, while a boundary layer-based sliding mode control term with a direct adaptive gain was designed to compensate for the unknown time-varying external disturbances. A Lyapunov analysis was carried out to formally prove the stability and tracking convergence of the proposed controller and derive the adaptation update laws.\looseness=-1

Simulation experiments on a full-cable model demonstrate that compared to a baseline controller, the proposed cable estimation-based adaptive robust control design results in reliable tracking performance and adaptive system identification, enabling the underactuated brachiating
robot to traverse the entire length of the cable in a continuous fashion in the presence of parametric uncertainties, bounded unstructured disturbances, and actuator saturation.

\bibliographystyle{IEEEtran}
\bibliography{cdc-bibtex}

\end{document}